\documentclass[11pt]{article}
\usepackage[preprint]{acl}
\usepackage{times}
\usepackage{latexsym}
\usepackage[T1]{fontenc}
\usepackage[utf8]{inputenc}
\usepackage{microtype}
\usepackage{inconsolata}
\usepackage{graphicx}
\usepackage{booktabs}
\usepackage{array}
\usepackage{placeins}
\usepackage{amsmath}
\usepackage{algorithm}
\usepackage{algpseudocode}
\usepackage{xurl}
\usepackage[skins]{tcolorbox}
\definecolor{choiceink}{HTML}{26324D}
\definecolor{choicepaper}{HTML}{F7F7F9}
\tcbset{choicecallout/.style={
  enhanced, colback=choicepaper, colframe=choiceink,
  boxrule=0.45pt, arc=1.2pt,
  left=4pt, right=4pt, top=6.5pt, bottom=2pt,
  before skip=8pt, after skip=6pt,
  fontupper=\normalsize, fonttitle=\normalsize\bfseries,
  coltitle=white,
  attach boxed title to top left={xshift=5pt,yshift=-7pt},
  boxed title style={colback=choiceink,colframe=choiceink,
    boxrule=0pt,arc=1pt,
    left=4pt,right=4pt,top=2pt,bottom=2pt}
}}
\definecolor{promptblue}{HTML}{2463A0}
\newcommand{\promptvar}[1]{\textcolor{promptblue}{\texttt{\{#1\}}}}
\newtcolorbox{promptbox}[1]{enhanced,colback=choicepaper,colframe=choiceink,
  title={#1},fonttitle=\small\bfseries,fontupper=\small,
  colbacktitle=choiceink,coltitle=white,boxrule=.45pt,arc=1pt,
  left=5pt,right=5pt,top=4pt,bottom=4pt,before skip=5pt,after skip=5pt}
\newtcolorbox{researchquestionbox}{choicecallout,title=Research Question}
\newtcolorbox{keyfindingsbox}{choicecallout,title=Key Findings}
\newtcolorbox{takeawaybox}{choicecallout,title=Takeaway}
\providecommand{\tightlist}{\setlength{\itemsep}{0pt}\setlength{\parskip}{0pt}}
\title{Correct Choices, Failed Sequences: Choice2Seq Benchmarks Correctness Retention Across Output Interfaces}
\author{
Tianyou Lai$^{1}$\thanks{These authors contributed equally to this work.} \quad
Wentao Yue$^{2}$\footnotemark[1] \quad
Zhiying Deng$^{3}$ \quad
Qilei Li$^{3}$\thanks{Corresponding author: \texttt{qilei.li@ccnu.edu.cn}.} \\
{\normalfont $^{1}$Carnegie Mellon University} \\
{\normalfont $^{2}$Shenzhen University} \\
{\normalfont $^{3}$Central China Normal University}
}
\hypersetup{
  pdftitle={Correct Choices, Failed Sequences: Choice2Seq Benchmarks Correctness Retention Across Output Interfaces},
  pdfauthor={Tianyou Lai, Wentao Yue, Zhiying Deng, Qilei Li}
}
\begin{document}
\maketitle
\begin{abstract}
\frenchspacing
A model may select the correct answer yet fail to construct it through a sequence of local decisions.
We introduce Choice2Seq, a diagnostic benchmark that measures correctness retention across complete-answer selection and sequential construction in independent runs on the same problems and candidate sets.
Sequence Composition Rate (SCR) measures the fraction of correctly selected items that are also answered correctly through local construction.
Jev achieves 97.1\% and 99.5\% selection accuracy on arithmetic and lexical retrieval, yet retains only 33.0\% and 2.1\% of those correct choices under serial construction, respectively.
Substantial gaps also emerge for StartLux and Qwen, including when Qwen emits characters directly.
Restricting local options to candidate-supported characters improves construction, while explicit position indexing improves copying fidelity, even for incorrect selections.
These interventions distinguish task correctness, plan fidelity, and output compliance.
Matched dialogue-state evaluations reveal that local field construction largely preserves Jev's correct states but reduces Qwen's mean-turn accuracy, highlighting task- and interface-dependent differences.
Sequence-level confidence and request-cost analyses further characterize completed outputs. Code and data will be released soon.
\end{abstract}
\begin{figure*}[!t]
\centering
\includegraphics[width=\textwidth]{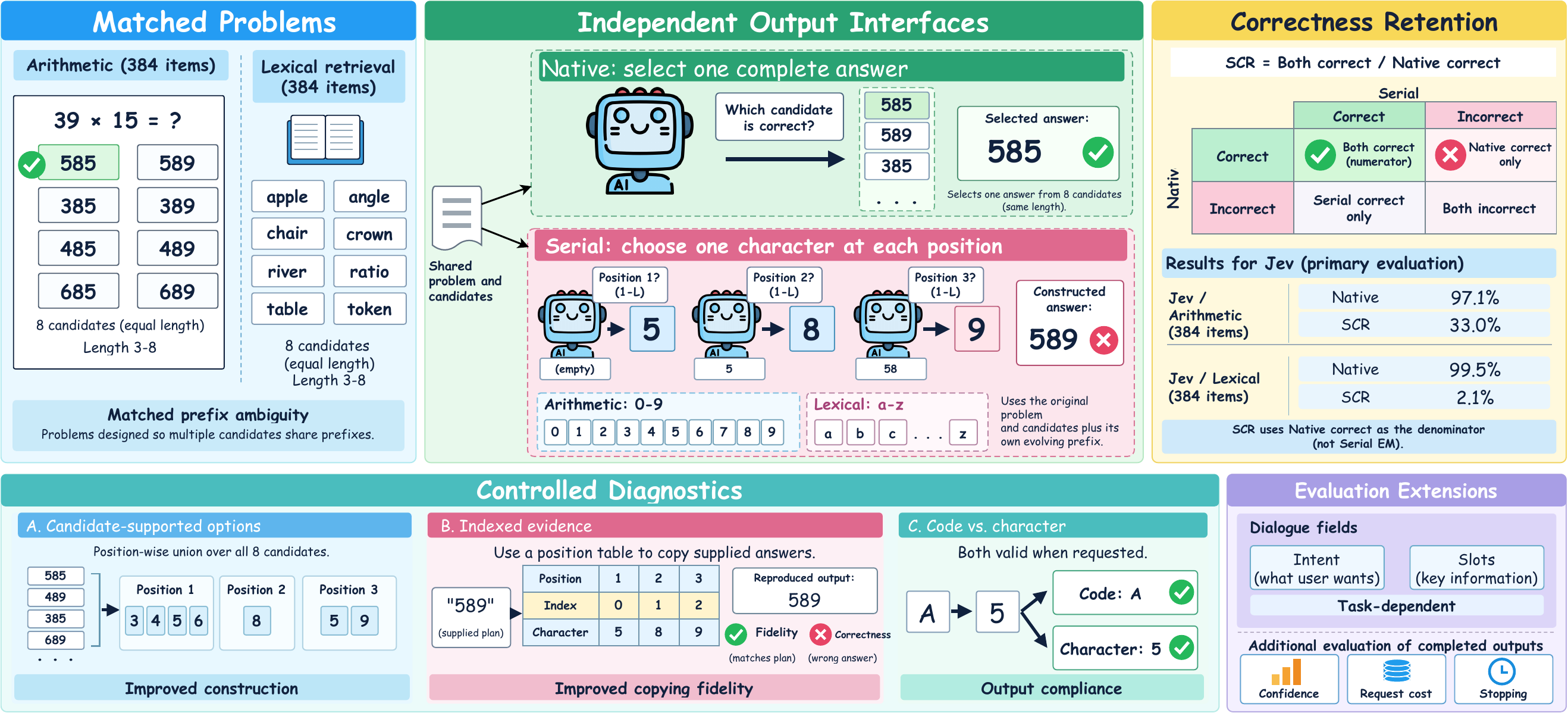}
\caption{
Choice2Seq distinguishes correctness retention, copying fidelity,
and output compliance. Upper: independent interfaces on matched
problems: correct complete-answer selection, incorrect Serial
construction. Lower: controls test support, evidence access,
rendering; extensions assess dialogue-state and completed-output
reliability.
}
\label{fig:overview}
\end{figure*}
\section{Introduction}\label{introduction}

\begin{quote}
\small
\emph{``The reductionist hypothesis does not by any means
imply a `constructionist' one.''}

\hfill --- Philip W. Anderson, \emph{More Is Different} (1972)
\end{quote}

Correctness is observed through an output interface, not in
isolation. Choosing among complete candidate answers tests a
single decision, whereas constructing an answer character by
character requires successive local decisions at each output
position, conditioned on the model's own prefix.
This distinction is especially relevant to Jev, which
exposes finite, typed decisions \citep{almeida2026introducing}
and has documented difficulties with repeated character
choices \citep{typesafe2026jaggedness}.
Strong accuracy on individual decisions alone therefore does not
establish correctness of the completed sequence.

\begin{researchquestionbox}
\textbf{How reliably does correctness in complete-answer
selection carry over to sequential construction on the same
problems, and how does it vary across interface conditions?}
\end{researchquestionbox}

Previous Jev studies examine decision accuracy, selective
judgment, option-name sensitivity, and probability coherence
\citep{deusser2026evaluating, li2026judge, sun2026type,
li2026beyond}. Yet aggregate accuracy across interfaces
obscures the key quantity: which correctly selected items
remain correct when constructed sequentially.
Interpreting the gap is also difficult because the interfaces
change decision granularity, option sets, and context together.
Furthermore, task correctness must be distinguished from copying
fidelity and output compliance: faithfully reproducing a
selected answer can still preserve an error.
These challenges require item-level pairing and targeted
controlled comparisons with distinct evaluation criteria.

To address these challenges, we introduce
Choice2Seq
(Figure~\ref{fig:overview}),
which separates measuring correctness retention from diagnosing
its limits. Native selects a candidate, whereas Serial
constructs an answer through local decisions
conditioned on its own prefix. Both run independently on the
same problems and candidate sets. Sequence Composition Rate
(SCR) measures the fraction of Native-correct items also solved
by Serial, using arithmetic and lexical targets with
matched candidate ambiguity along correct prefixes.
Rather than attributing an accuracy gap to a mechanism,
we vary local choice support and context, and test reproduction
of supplied answers, including incorrect selections, under
different evidence and output formats. These diagnostics
distinguish task correctness, copying fidelity, and format
compliance.

Our paired results reveal a large but nonuniform retention gap.
Jev's Native accuracy reaches 97.1\% on arithmetic and 99.5\%
on lexical retrieval, yet SCR is only 33.0\%
and 2.1\% (Table~\ref{tab:primary}).
Similar losses appear for StartLux and persist for Qwen under
direct character output, showing that typed response codes
alone cannot explain the gap.
Candidate-supported options improve construction, whereas
indexed evidence improves copying fidelity even for incorrect
selected answers (Figure~\ref{fig:candidate-support};
Table~\ref{tab:recovery}).
Matched symbolic tasks and dialogue fields show
that poor retention is not universal: Jev largely preserves
correct dialogue states, while Qwen loses mean-turn accuracy
(Figure~\ref{fig:workflow}).

Our contributions are:

\begin{itemize}
\tightlist
\item
  We introduce Choice2Seq and SCR to quantify item-level
  correctness retention across independently executed output
  interfaces, beyond aggregate accuracy comparisons.
\item
  We show that task correctness, copying fidelity, and output
  compliance require separate evaluation, using targeted
  controls over local choices, input evidence, and output formats.
\item
  We establish task- and model-dependent boundaries of
  retention through symbolic and dialogue evaluations, and
  assess confidence and request costs for completed outputs.
\end{itemize}

\begin{keyfindingsbox}
\begin{list}{\textbullet}{\setlength{\leftmargin}{1em}
  \setlength{\labelsep}{0.4em}\setlength{\itemsep}{4pt}
  \setlength{\parsep}{0pt}\setlength{\topsep}{0pt}}
\item \textbf{High selection accuracy does not ensure
correctness retention.} Large Native--Serial gaps persist
across models, including when Qwen emits characters directly
(Table~\ref{tab:primary}).
\item \textbf{Faithful construction need not yield correct
answers.} Candidate-supported choices improve task accuracy,
while indexed evidence improves copying fidelity, even for
incorrect selections (Figure~\ref{fig:candidate-support}
and Table~\ref{tab:recovery}).
\item \textbf{Sequential construction is not inherently
unreliable.} Matched symbolic tasks demonstrate high retention;
in dialogue states, Jev largely preserves correct predictions
while Qwen loses mean-turn accuracy
(Figure~\ref{fig:workflow}).
\end{list}
\end{keyfindingsbox}
\section{Related Work}\label{related-work}

\textbf{Structured decisions and output interfaces.}
Jev produces finite, typed choices with associated probabilities
\citep{almeida2026introducing}.
Decision-model studies examine general capabilities, security specialization,
confidence-based escalation, legal judgment, and downstream
scientific decisions
\citep{deusser2026evaluating,chen2026secjev,li2026judge,
zhang2026same,deng2026scientific}.
Output constraints also affect generative-model accuracy
\citep{tam2024speak}, while JSONSchemaBench evaluates constraint
coverage, output quality, and efficiency
\citep{geng2025jsonschemabench}.
Choice2Seq pairs complete-answer and local outcomes on identical
problems and candidates, identifying retained, lost and newly correct
items. Matched fields in Schema-Guided Dialogues extend this
comparison beyond enumerated complete answers
\citep{rastogi2020towards}.

\textbf{Construction and readout.}
Multiple-choice selection is sensitive to labels and
positions \citep{zheng2024selectors}, while token-based scoring
can disagree with generated answers \citep{wang2024look}.
Jev's decisions vary with option names and answer-preserving
context \citep{sun2026type,xu2026jevout}, paralleling
prompt-format sensitivity \citep{sclar2024quantifying}.
Spelling research distinguishes character information encoded
in model representations from reliable use during generation
\citep{itzhak-levy-2022-models,edman-etal-2024-cute,
hiraoka-inui-2025-spelling}.
Jev's indirection sensitivity raises a
concern \citep{typesafe2026jaggedness}.
Exposure-bias studies examine error accumulation and
recovery under generated prefixes
\citep{he2019exposure,arora2022exposure}.
Unlike continuous decoding, Choice2Seq probes prefix feedback
through fresh position queries with external state, alongside
controls for choice support, evidence access, and output contracts.
Supplied-answer controls separate copying fidelity from task
correctness, and matched digit--letter operations test
representation dependence.

\textbf{Confidence and trajectory outcomes.}
Calibration depends on the correctness event being scored
\citep{guo2017calibration,jagannatha2020calibrating,jiang2021know}.
Uncertainty research distinguishes token-, sequence-, and
semantic-level assessments
\citep{malinin2021uncertainty,kuhn2023semantic},
while Jev coherence tests examine consistency across related
requests \citep{li2026beyond}.
Metric effectiveness varies across output contexts
\citep{deviyani2025contextual}.
Choice2Seq tests whether scores aggregated from successive local
choices predict completed-string exact match, comparing ranking with
length-only references and calibration with prevalence baselines.
Trajectory-aware evaluation captures intermediate outcomes
beyond terminal success \citep{diaz2026offline}.
Our dialogue analyses therefore report mean-turn, final-state,
and trajectory accuracy alongside replay-based state recovery.

\section{Evaluation Protocol}\label{evaluation-protocol}

Choice2Seq compares correctness across whole-answer selection
and local construction in character tasks and semantic fields.
Paired outcomes reveal item-level correctness turnover;
interventions identify interface conditions that preserve correctness.

\subsection{Interfaces and outcomes}\label{interfaces-and-outcomes}

Each item contains a problem \(x_i\), target \(y_i\) of known length
\(L_i\), and eight distinct, equal-length candidates \(A_i\).
Native selects a complete candidate. Independent queries each position
without a prefix. Serial starts empty and appends one character per
position. Each step freshly requests the original answer's character
at an absolute position, receiving the problem, candidates, length and
current prefix as external state. The prefix is a fallible draft. Local options
are ten digits for arithmetic or lowercase letters for lexical retrieval.
Algorithm~\ref{alg:paired-evaluation} in
Appendix~\ref{app:protocol} gives the complete procedure.

Let \(N_i\) indicate a correct Native choice and \(S_i\) an entirely
correct Serial string. Sequence Composition Rate (SCR) measures the
conditional correctness of Serial on the Native-correct items:
\begin{equation}
\mathrm{SCR}=\frac{\sum_i \mathbf{1}[N_i=S_i=1]}
                       {\sum_i \mathbf{1}[N_i=1]}.
\label{eq:scr}
\end{equation}
The paired table records both correct, Native only, Serial only and both
wrong. Marginal EM combines losses and gains, which can cancel or mask
replacement of successful items. SCR summarizes the overlap within
Native successes; Serial success on Native-wrong items records gains
outside that population. This distinction reveals turnover even when
Serial EM improves (Table~\ref{tab:turnover-main}). The runs are independent:
neither receives the other's answer. For Native EM $a$ and Serial EM $b$,
$|\mathrm{SCR}-b|\leq 1-a$, so near-perfect Native accuracy makes the two
rates numerically close. We report the joint counts, Native coverage,
SCR numerator and denominator, and gain rate alongside Serial content EM
on all items. Loss mass is the Native-only count divided by all items
(Appendix~\ref{paired-outcome-interpretation}).

Fidelity instead compares construction with an answer supplied to it.
TeacherForced supplies gold prefixes; gold Copy supplies the complete
target; SelfPlanSpell and fresh-plan copying supply the model's selected
answer, including incorrect selections. Gold EM scores the original task,
while fidelity scores copying. Program return emits the selected string
without another request, preserving Native accuracy. A common-parent
comparison appears in Appendix~\ref{same-parent-choice-workflows}.
Independent and Serial execute $L_i$ positions in the content track;
a separate common alphabet
adds punctuation, whitespace and END to evaluate termination
(Appendix~\ref{app:protocol}).

\subsection{Tasks and candidate geometry}\label{tasks-and-controlled-interventions}

The formal test contains 384 items per domain, balanced over lengths
three to eight, with disjoint development data. Multiplication crosses
operand-length balance and carry complexity, the latter computed by
a column-arithmetic simulator. Lexical items retrieve English nouns
from WordNet definitions with wordfreq metadata
\citep{miller1995wordnet,wordnet2006resource,speer2022wordfreq}.
Screening and human review establish sense-specific acceptability within
the displayed candidates, excluding target cues, same-synset distractors
and semantic conflicts. Strict EM requires an exact match to the
target; an output audit distinguishes candidate membership,
WordNet indexing and blinded semantic judgments on reviewed outputs
(Appendix~\ref{lexical-output-audit}). Frozen frequency and repetition
quotas govern sampling; the independent confidence cohort uses separate
targets and length quotas determined by vocabulary availability
(Appendix~\ref{app:protocol}).

Candidate geometry determines how much information a prefix supplies.
For a correct prefix of length \(k\), \(K_i(k)\) counts candidates
consistent with that prefix. Each arithmetic item is paired with
a lexical item sharing the full vector of correct-prefix counts,
including \(K_i(0)=8\) and \(K_i(1)=2\).
We also record prefix uniqueness and entropy under uniform
candidate probabilities. This matches ambiguity along the
reference path, while off-path branching and position-wise
character support remain task-specific.
This matching does not equalize computation or lexical retrieval difficulty.
A Native control receives only the candidates to measure
selection from candidate-set cues.
An isomorphic COPY/REVERSE panel uses bijective digit/letter
encodings to match source structure, operation, ten-symbol
inventory, and candidate geometry.
A semantic bridge constructs named Boolean fields or binary
outputs from the same source facts.

\subsection{Controlled interventions}\label{controlled-interventions}

The interventions vary feasible outputs, evidence access and rendering.
Full Serial allows the full domain alphabet at every position.
Union restricts local choices to characters appearing at that
position among the candidates, permitting cross-candidate
recombinations. Trie restricts each choice to continuations
compatible with the generated prefix and candidate set, so
every valid completed string belongs to the original eight candidates.
Support filtering preserves option identities and relative order.

A matched panel crosses these supports with controller filling.
On valid trajectories, Call variants query every position, including singletons; Fill variants
supply singleton values without a request. FullFill supplies only
position-union singletons and keeps the full alphabet elsewhere. These
contrasts distinguish output-space restriction from saved requests;
ambiguous decision counts accompany both. Contemporary Native and Full
repeats use all original items for Jev and a frozen, stratified 192-item
Qwen3.5-27B subset. Original Union results retain their separate panel
(Appendix~\ref{matched-feasible-output-control}).

Raw supplies a string; Indexed adds its position table; Direct also
supplies the queried character. Fixed-criterion copying holds questions,
options and order constant across evidence and prefix conditions.
Padding and Batch controls test length and bundling; selected and
incorrect targets distinguish fidelity from correctness. Rebinding and
matched code/character contracts test binding and rendering. Paired samples, prompt
variants and inference families appear in
Appendices~\ref{app:protocol} and~\ref{app:revision}.

Schema-Guided Dialogues \citep{rastogi2020towards} extend the comparison
to known-service intents and categorical slots. With identical history
and schemas, Joint requests all fields in one call; Independent requests each
field separately; Serial also receives its own earlier predictions for the current
turn. Local arms receive neither complete-state candidates nor Joint's
answer. A separate replay compares transcript recomputation with
predicted or gold prior state. Endpoints include dialogue-averaged
mean-turn EM, final-state and trajectory accuracy, and request costs
(Appendix~\ref{app:extensions}).

\subsection{Models and statistical analysis}\label{models-and-statistical-analysis}

Jev's formal construction uses choices returned by the service.
Readout and plan-copying diagnostics instead use a predefined
argmax over reported probabilities. Both selections and reported
probabilities are retained. StartLux uses pinned BF16 weights
and released decision code with supplied names and descriptions
\citep{startlux2026release}. Qwen3-8B and Qwen3.5-27B use full BF16
weights with thinking disabled \citep{qwen2025model,qwen2026model35}.
Queries for legal codes verify allowable code tokens, while direct
character and complete-string responses use greedy generation.
Serial measures external sequential construction: each new request
reconstructs state and asks an absolute-position question.
WholeGenerate solves the same Qwen subset in one uninterrupted response.
PrefixContinue appends the current prefix to the assistant prompt
and consumes the first character of each generated suffix, without
an absolute-position query. Different output contracts retain separate
Native denominators. Invalid responses fail strict scoring.
Complete prompts, decoding budgets and runtime identities appear
in Appendices~\ref{app:protocol},
\ref{app:extensions} and~\ref{app:prompts}.

An independence reference multiplies observed position accuracies within
model, domain and length. Comparing Independent/TeacherForced
string EM with this reference captures deviations potentially arising
from item heterogeneity or dependent errors; Serial
additionally varies prefix context. Whole-item bootstrap preserves
heterogeneity and correlated errors
(Appendix~\ref{empirical-composition-reference}).

Intervals resample whole items or source clusters within frozen strata,
redrawing conditional denominators and keeping repeated measurements
together \citep{dror2018hitchhikers}. Primary exact McNemar tests use
Holm correction \citep{mcnemar1947note,holm1979simple}; supplementary
panels specify their own families. Dialogue bootstrap retains whole
dialogues and checks groups linked by shared responses. Exploratory GEE
clusters by item and uses only converged fits with finite estimates
and no separation \citep{liang1986longitudinal}.
Zero-success SCR cells use exact two-sided binomial intervals; other
main-table cells use the item bootstrap. Appendices give seeds and prompts.
Prompt robustness reports individual settings, mean and worst performance
across variants, and common-parent ranking stability.
\section{Correctness Retention Across Interfaces}
\label{correctness-retention-across-interfaces}

We first quantify the paired interface gap across tasks and model
families, then examine how controlled interventions affect correctness
retention.

\begin{table*}[!t]
\centering
\small
\caption{Paired correctness on 384 items per domain. Joint counts give both correct / Native only / Serial only / neither; EM and SCR are percentages. SCR counts are both correct / Native correct. Intervals are length-stratified item bootstrap 95\% CIs, except zero-success cells (\textdagger{}), which show exact two-sided binomial intervals.}
\label{tab:primary}
\setlength{\tabcolsep}{3pt}
\begin{tabular}{@{}llrrrrr@{}}
\toprule
Model / interface & Domain & Native EM & Joint counts & Serial EM & SCR count & SCR {[}95\% CI{]} \\
\midrule
Jev & math & 97.1 & 123/250/0/11 & 32.0 & 123/373 & 33.0 {[}29.0, 37.0{]} \\
Jev & lexical & 99.5 & 8/374/0/2 & 2.1 & 8/382 & 2.1 {[}0.8, 3.7{]} \\
StartLux & math & 61.2 & 8/227/1/148 & 2.3 & 8/235 & 3.4 {[}1.3, 5.9{]} \\
StartLux & lexical & 93.8 & 1/359/0/24 & 0.3 & 1/360 & 0.3 {[}0.0, 0.8{]} \\
Qwen3-8B (code) & math & 15.4 & 0/59/0/325 & 0.0 & 0/59 & 0.0 {[}0.0, 6.1{]}\textdagger{} \\
Qwen3-8B (code) & lexical & 94.5 & 0/363/0/21 & 0.0 & 0/363 & 0.0 {[}0.0, 1.0{]}\textdagger{} \\
Qwen3.5-27B (code) & math & 25.3 & 1/96/1/286 & 0.5 & 1/97 & 1.0 {[}0.0, 3.4{]} \\
Qwen3.5-27B (code) & lexical & 99.7 & 5/378/0/1 & 1.3 & 5/383 & 1.3 {[}0.3, 2.6{]} \\
Qwen3.5-27B (direct) & math & 98.4 & 39/339/0/6 & 10.2 & 39/378 & 10.3 {[}7.4, 13.3{]} \\
Qwen3.5-27B (direct) & lexical & 99.0 & 132/248/0/4 & 34.4 & 132/380 & 34.7 {[}30.0, 39.6{]} \\
\bottomrule
\end{tabular}
\end{table*}

\begin{figure*}[t]
\centering
\includegraphics[width=.91\textwidth]{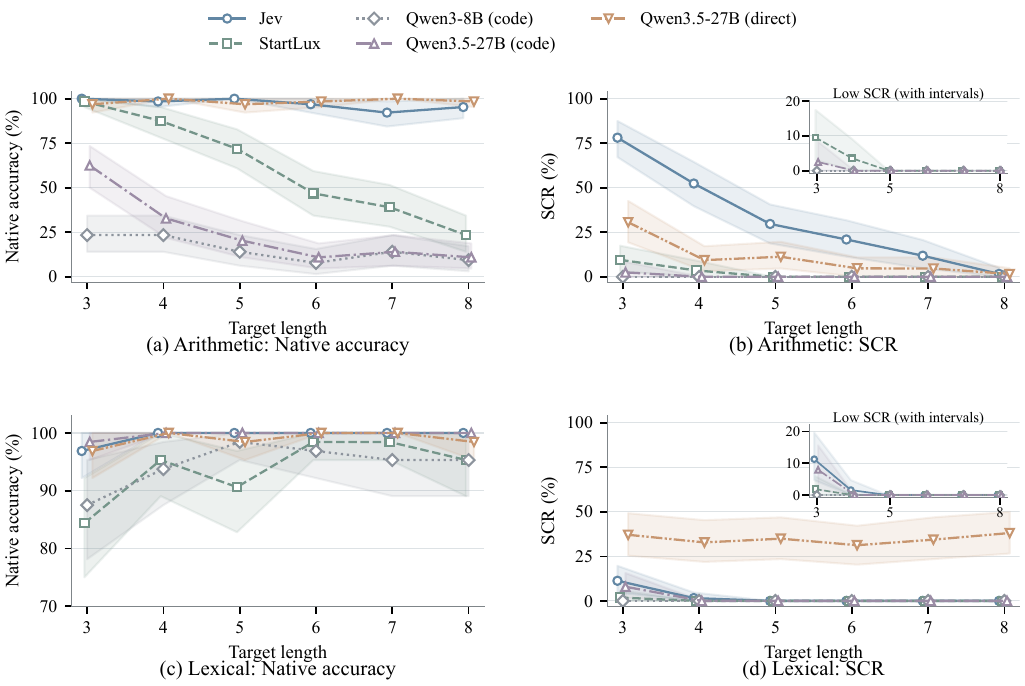}
\caption{High Native accuracy does not guarantee Serial retention.
Across 64 paired items per domain and length, (a,c) show near-ceiling
Native accuracy for Jev and direct-output Qwen3.5-27B.
In (b,d), Jev's arithmetic retention declines with length while lexical
retention remains near zero; direct Qwen instead retains more lexical
than arithmetic answers. SCR conditions on Native-correct items;
bands indicate 95\% item-bootstrap intervals.}
\label{fig:length}
\end{figure*}

\textbf{High whole-answer accuracy coexists with low local construction
accuracy.} Jev excels in Native but retains fewer correct
answers under Serial, especially in lexical retrieval
(Table~\ref{tab:primary}). Every original paired decrease survives
Holm correction. Candidates alone yield near-chance selection,
supporting a role for problem evidence. Qwen's code and
character contracts yield different Native-correct populations;
shared-correct comparisons preserve the Jev/Qwen domain reversal
(Appendix~\ref{full-population-reconstruction}). Native ranks direct Qwen
above Jev on arithmetic, while Serial reverses that ordering.

\textbf{Marginal improvement can replace previously correct answers.}
High Native accuracy makes SCR close to Serial EM in the natural tasks,
with the same model ordering. Named semantic fields reveal correctness
turnover (Table~\ref{tab:turnover-main}): Qwen3-8B gains nine successes
overall but retains six of 28 Native successes, losing 22 and gaining 31.
SCR distinguishes preserved successes from their replacement; the joint
table records both. Jev's similar marginal accuracies accompany high
overlap. Complete transitions appear in
Appendix~\ref{paired-correctness-turnover}.

\begin{table}[!t]
\centering\small
\caption{A marginal gain can replace most Native successes. Named semantic fields use 128 records per model. Joint counts are both correct / Native only / Serial only / neither; EM and SCR are percentages.}
\label{tab:turnover-main}
\setlength{\tabcolsep}{3pt}
\begin{tabular}{@{}lrrr@{}}
\toprule
Model & Native/Serial EM & Joint counts & SCR\\
\midrule
Jev & 89.8/90.6 & 111/4/5/8 & 96.5\\
Qwen3-8B & 21.9/28.9 & 6/22/31/69 & 21.4\\
StartLux & 51.6/71.1 & 51/15/40/22 & 77.3\\
\bottomrule
\end{tabular}
\end{table}

\textbf{Continuous generation can succeed where position queries fail.}
On the same Qwen3.5-27B checkpoint and 96 parents per domain,
WholeGenerate solves almost all items in one response, whereas Serial
position queries lose many successes despite valid character output
(Table~\ref{tab:interface-generation}). PrefixContinue reconstructs the
assistant prefix on every request: arithmetic remains accurate, while
lexical trajectories develop both content and format failures. This controller
is distinct from uninterrupted decoding. All 1,056 Serial position requests
finish within two output tokens, below their eight-token cap.
The contrast demonstrates failure under the evaluated external construction
interface; prompt-state representation, assistant prefill and rendering
also vary (Appendices~\ref{natural-generation-interface-control}
and~\ref{generation-budget-audit}).

\begin{table}[!t]
\centering\small
\caption{One-response solving largely preserves accuracy; repeated position queries do not. Qwen3.5-27B uses the same 96 items per domain, with no supplied answer. Correct counts are content EM; valid trajectories are arithmetic/lexical percentages; calls are per item.}
\label{tab:interface-generation}
\setlength{\tabcolsep}{3pt}
\begin{tabular}{@{}lrrrr@{}}
\toprule
Interface & Arithmetic & Lexical & Valid & Calls\\
\midrule
Native & 94/96 & 96/96 & 100/100 & 1.00\\
WholeGenerate & 94/96 & 91/96 & 100/100 & 1.00\\
Position query & 14/96 & 27/96 & 100/100 & 5.50\\
PrefixContinue & 94/96 & 0/96 & 100/9.4 & 5.50\\
\bottomrule
\end{tabular}
\end{table}

\textbf{The gap persists across tested prompts, with interface-dependent
magnitude.} Jev arithmetic retention decreases with length; lexical
retention is low even for short targets (Figure~\ref{fig:length}).
Prompt, order and repeat perturbations show the same broad Jev pattern,
though paired intervals are wide. Across three templates, direct Qwen Serial
EM ranges from 7.3--28.1\%
in arithmetic and 39.6--56.2\% in lexical retrieval; Native accuracy remains higher
in both domains throughout. The v2 improvements in both domains survive
Holm correction, and all responses remain format-valid
(Table~\ref{tab:revision28}). Mean and worst-template SCR and common-parent
ranking stability appear in Table~\ref{tab:prompt-aggregate} and
Appendix~\ref{observed-prompt-setting-summaries}.

\begin{takeawaybox}
\textbf{Compare which items are correct as well as how many.}
Paired outcomes distinguish preserved successes from losses and gains
that marginal accuracy can otherwise obscure.
\end{takeawaybox}

\section{Diagnosing Correctness and Fidelity}
\label{diagnosing-correctness-and-fidelity}

Controls distinguish solving the original task, faithfully copying
supplied answers, and correctly rendering outputs under different
response contracts.

\subsection{Candidate support and prefix context}
\label{candidate-support-and-evidence-access}

\textbf{Candidate-consistent support substantially improves lexical construction.}
Union permits unlisted character recombinations; TrieCall preserves Native's
eight feasible complete outputs while still querying every singleton choice.
On matched Jev lexical items, EM rises from 1.8\% under Full to
24.0\% under UnionCall and 82.0\% under TrieCall
(Figure~\ref{fig:matched-feasible-outputs}). Only 26.3\% of UnionCall
outputs belong to the candidates; every valid Trie output does. Trie
improves over Full in both models and domains, and over lexical Union
after correction. Jev's domain contrast shrinks to 2.6 points
[${-}2.9$, 7.8], although Native remains more accurate.

The observed lexical recovery is consistent with differences in candidate
geometry. Matching correct-prefix counts leaves arithmetic unions with
eight feasible strings, but lexical unions admit many cross-candidate
recombinations (Table~\ref{tab:revision23}). TrieCall prevents these
recombinations while retaining every position query. A wrong Trie branch
excludes gold, so recovery still requires selecting the right candidate
through local decisions.

\textbf{Restricted support and saved requests have different effects.}
FullFill improves arithmetic accuracy while keeping full choices at ambiguous
positions. Union and Trie restrict those choices; Fill additionally skips
singletons. TrieCall still queries all positions; TrieFill reduces calls from
5.5 to about two per item while keeping the same eight feasible outputs.
In this panel, lexical Union has no singleton positions and queries every
position. On twelve-character reversals, queried depth and
branch placement affect Jev more than alphabet size. Complete comparisons
retain decision counts and feasible outputs
(Appendices~\ref{matched-feasible-output-control}
and~\ref{controller-factor-study}).

\textbf{Local improvement need not recover complete strings.}
Generated prefixes help Jev arithmetic over Independent queries;
both lexical outcomes remain low. Gold prefixes improve item-averaged
character accuracy, especially lexically, without comparable gains in
complete-string success. The paired arithmetic TeacherForced string-success
difference has an interval spanning zero (Appendix~\ref{app:results}).
The product of Independent position accuracies yields a low
lexical EM reference. Arithmetic Serial exceeds this reference at most lengths,
consistent with useful prefix feedback (Figure~\ref{fig:composition-reference}).
Deviations from this reference may reflect dependent errors and item
heterogeneity; the Serial contrast also changes prefix context.
Together, these controls distinguish local error accumulation from
additional interface effects. Imperfect position decisions already
predict low string EM; prefix feedback can change those decisions;
and candidate consistency determines which complete strings they can
form. Trie recovery and accurate WholeGenerate outputs demonstrate
that lexical losses depend on how the external controller exposes and
combines choices, rather than representing failures shared by all
sequence interfaces.

\subsection{Evidence access and plan fidelity}
\label{faithful-copying-and-task-correctness}

\textbf{An available answer still needs a usable representation.}
Raw gold Copy and SelfPlanSpell fail on many strings, especially lexical
targets. A position table derived from the supplied answer improves
both position readout and complete-string Serial fidelity for Jev and StartLux
in every fixed-criterion model/domain comparison. Questions, options
and order remain constant while evidence conditions are crossed with
Independent and Serial querying (Figure~\ref{fig:copy-controls}).
Indexing simultaneously changes addressability, salience and prompt length.

\begin{table}[tbp]
\centering
\small
\caption{Serial plan-copying fidelity and indexed task correctness. Raw and Indexed measure fidelity to freshly selected model plans, including incorrect selections; Gold EM measures indexed outputs against the task target. Each model--domain pair contains 30 plans.}
\label{tab:recovery}
\setlength{\tabcolsep}{3pt}
\begin{tabular}{@{}llrrr@{}}
\toprule
Model & Domain & Raw & Indexed & Gold EM \\
\midrule
Jev & math & 17/30 & 30/30 & 28/30 \\
Jev & lexical & 4/30 & 29/30 & 29/30 \\
StartLux & math & 8/30 & 25/30 & 16/30 \\
StartLux & lexical & 1/30 & 29/30 & 28/30 \\
\bottomrule
\end{tabular}
\end{table}

\textbf{Better copying can preserve an incorrect answer.}
Fresh-plan indexing uses each model's selected string, including
incorrect selections. It improves fidelity for both models and domains
under paired correction (Table~\ref{tab:recovery}; Figure~\ref{fig:ownplan}),
but faithfully copied incorrect plans fail gold EM. On supplied
targets enriched with errors, Jev faithfully copies every indexed
incorrect target without solving the original problem
(Appendix~\ref{app:revision}). Fidelity and task correctness therefore
require separate scores.

\textbf{Structured evidence helps beyond increasing its length.}
Indexed copying outperforms nonindexed padding, including Qwen controls
matched on rendered input tokens (Table~\ref{tab:interface-padding}).
Information and salience still differ. Controller and Batch contrasts
appear in Appendix~\ref{fixed-criterion-copy-follow-up}; program return
provides the zero-additional-request reference.

\subsection{Option binding and output contracts}
\label{option-binding-and-output-contracts}

\textbf{Option names and output contracts affect readout.}
Remapping controls motivated by option-name conflicts
\citep{sun2026type} show corrected extra losses for StartLux arithmetic
and Qwen lexical readout, with none observed for Jev
(Appendix~\ref{app:results}). Matched Qwen generation favors character
output with indexing but codes with direct lexical evidence. In the
full prompt/order grid, Qwen3-8B has corrected advantages in both output
directions, whereas Qwen3.5's corrected differences favor codes
(Figure~\ref{fig:robustness}). Many strict Qwen3.5 character failures
instead return the uppercase code bound to the correct criterion.
Natural Serial retains content errors despite valid character output.
Complete contracts, paired results and request examples distinguish
these outcomes (Appendices~\ref{app:revision} and~\ref{app:prompts}).

\subsection{Task and representation effects}
\label{task-and-representation-boundaries}

\textbf{Task structure changes the retention pattern.}
Matched COPY/REVERSE operations yield a small letter-minus-digit Serial
contrast whose interval spans both directions, rather than reproducing
the natural lexical disadvantage. Jev also retains most correct named
and binary semantic fields (Table~\ref{tab:extended3}). The natural
arithmetic--lexical contrast therefore combines solving
with candidate geometry and evidence access. It narrows under Trie,
while direct Qwen reverses the domain ordering. The twelve-character
factor panel further identifies queried depth and branch placement as
contributors to construction accuracy. Together, these controls associate retention with task structure and
the output interface, beyond digit-versus-letter labels.

\begin{takeawaybox}
\textbf{Choose the evaluation endpoint before judging any intervention.}
Evaluate solving with task accuracy, copying with plan fidelity, and
rendering with output-contract compliance.
\end{takeawaybox}

\section{Reliability of Completed Outputs}
\label{evaluating-constructed-outputs}

\subsection{Confidence, cost, and stopping}
\label{sequence-confidence-cost-and-stopping}

\textbf{Confidence must be assessed against the completed object.}
We score chosen-path product, geometric mean and minimum top-choice
probability using Brier, ECE, AUROC and average precision (AP)
\citep{brier1950verification,guo2017calibration,davis2006relationship}.
Product improves arithmetic ranking over the length baseline.
Calibration error persists. Independent lexical evaluation yields 13/2,347
successes (0.55\%; exact 95\% interval 0.30--0.95\%). With so few
successes, AUROC/AP gains over length and the Brier difference from a
prevalence-only baseline have intervals spanning zero. At 10\% coverage,
Product precision is only 3.8\% [1.7, 6.4]. Successes concentrate at short
lengths; common-length weighting changes the population, not the success
count (Appendix~\ref{app:revision}).

\textbf{Cost and termination require separate endpoints.}
Serial adds requests, tokens and recorded request time relative to
Native plus program return. The common-alphabet track also separately
scores premature/missing END and cap exhaustion alongside strict content
EM (Appendix~\ref{app:results}).

\subsection{Field construction and downstream state}
\label{field-construction-and-downstream-state}

\textbf{Field construction tests granularity without complete-state
candidates.} Under Serial, Jev retains 420 of 429 Joint-correct turns;
its mean-turn difference interval spans zero. Qwen shows a corrected
Serial mean-turn loss, while intervals for Serial--Joint final-state
differences span zero for both models (Figure~\ref{fig:workflow}).
Local construction assembles fields without Joint's selected state
and incurs higher per-turn model request costs
(Appendix~\ref{app:extensions}).

\textbf{Trajectory EM can hide differences in later states.}
Oracle/Rollout tests gold versus predicted prior state: Rollout has
lower mean-turn and final-state accuracy despite identical trajectory
EM, which stays zero after the first error
(Appendix~\ref{app:extensions}).
Trajectory-aware evaluation similarly preserves progress beyond
terminal success \citep{diaz2026offline}.

\begin{takeawaybox}
\textbf{Score states after errors, not just error-free trajectories.}
Report later-state accuracy and recovery alongside trajectory EM,
and track construction costs separately for each interface.
\end{takeawaybox}

\section{Conclusion}\label{conclusion}

We asked whether correctness under whole-answer selection carries over
to independently executed local construction. We developed Choice2Seq
to compare interfaces on paired tasks and examine conditions
preserving correctness. High selection accuracy cannot guarantee
retention: candidate support, evidence representation and generation
protocols shape construction outcomes. Copying fidelity can improve
without improving task correctness, while later-state accuracy reveals
differences hidden by trajectory EM. These findings call for
interface-aware evaluation that tracks preserved successes and treats task correctness, copying fidelity and output
compliance as distinct endpoints.

\label{main:ended}
\section*{Limitations}\label{limitations}

Choice2Seq evaluates correctness retention under explicitly defined
output interfaces on the same underlying problem instances. The primary
tasks use known lengths of three to eight and eight displayed candidates,
covering arithmetic and English noun retrieval. A twelve-character
symbolic panel extends the length range under matched operations.
These settings characterize short, constrained construction rather
than unrestricted long-form or multilingual generation. Full and Union
can admit outputs beyond the candidate set, while Trie preserves
complete candidate membership through a different local decision
structure from Native. Matching correct-prefix geometry controls
candidate ambiguity along the reference path; off-path branching,
computation and vocabulary remain task-specific.

The natural-generation comparison evaluates continuous responses,
position queries and reconstructed character prefills as distinct
interface configurations. It does not attribute their differences to
a single internal decoding mechanism. Likewise, indexing controls
compare evidence representations, with addressability and salience
potentially changing together. The lexical audit classifies all
outputs in the matched panel against candidates and WordNet,
complemented by blinded semantic review of all indexed noncandidate
pairs and a stratified unindexed sample. Human judgments therefore
support the reviewed population, while resource-level conclusions
cover the full panel.

Dialogue experiments extend the framework to known-service intents
and categorical fields, using state-specific endpoints rather than
character SCR. Fresh-plan fidelity comparisons use 30 plans per
model/domain, with exact intervals quantifying finite-sample precision.
Confidence and cost findings apply to the evaluated length mixtures,
success prevalence, prompts and runtime configurations.

\section*{Ethical Considerations}\label{ethical-considerations}

We use generated arithmetic and synthetic facts, WordNet and wordfreq
with source notices, and public Schema-Guided Dialogues under CC BY-SA
4.0. Lexical resources can contain dated senses or terminology. The
release preserves study provenance and scientific response records while
excluding credentials, account identifiers and identifying paths. Human
review and automated lexical screening are documented separately.

\bibliography{references}
\clearpage
\appendix
\renewcommand{\dbltopfraction}{0.95}
\renewcommand{\dblfloatpagefraction}{0.95}
\renewcommand{\textfraction}{0.05}
\setcounter{dbltopnumber}{3}
\raggedbottom
\makeatletter
\setlength{\@fptop}{0pt}
\setlength{\@dblfptop}{0pt}
\setlength{\@fpsep}{\floatsep}
\setlength{\@dblfpsep}{\dblfloatsep}
\makeatother
\FloatBarrier
\setcounter{table}{0}
\renewcommand{\thetable}{A\arabic{table}}
\section{Detailed Protocol and
Reproducibility}
\label{app:protocol}
\begin{figure*}[t]
\centering
\includegraphics[width=\textwidth]{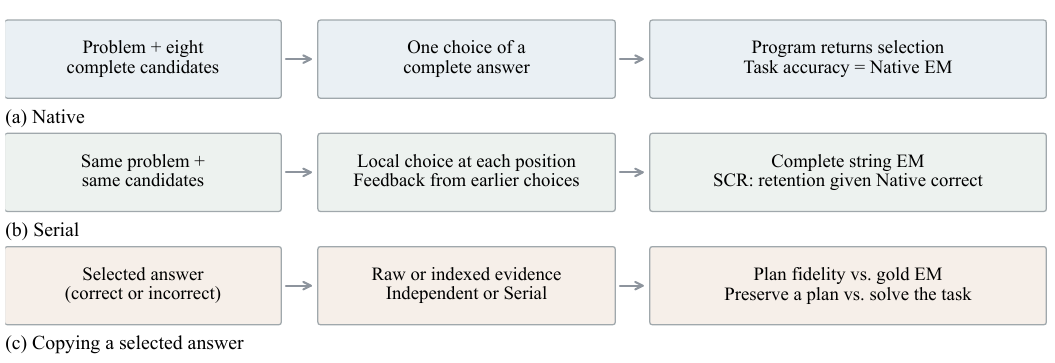}
\caption{Correctness retention and plan fidelity require different references. (a) Program return preserves a selected answer without another model request. (b) Serial solves the same problem separately, measuring overlap with Native correctness. (c) Copying scores fidelity to the supplied plan separately from gold-answer correctness, distinguishing faithful copying of a wrong plan from successful solving.}
\label{fig:protocol}
\end{figure*}\label{detailed-protocol-and-reproducibility}

\subsection{Design and independent
units}\label{design-and-independent-units}

The independent unit is a natural parent item, shared symbolic source,
semantic fact record or complete dialogue. Positions, conditions,
repetitions and option permutations remain attached to that unit.
Cross domain contrasts resample the designated arithmetic and lexical
partner pair together. Character construction probes the interface. The matched field task
extends evaluation to \mbox{states without complete-answer candidates.}

Algorithm~\ref{alg:paired-evaluation} gives the fixed length procedure
for Full, Union and prefix-aware Trie support. Native's selection is retained for paired
scoring. Each local call uses the independently constructed prefix.

\paragraph{Paired outcomes and SCR.}\label{paired-outcome-interpretation}
Let $p_{ab}$ denote the proportion with Native correctness $a$ and
Serial correctness $b$. Native EM is $p_{11}+p_{10}$, Serial EM is
$p_{11}+p_{01}$, and SCR is $p_{11}/(p_{11}+p_{10})$.
Loss and gain mass are $p_{10}$ and $p_{01}$; their difference is the
marginal accuracy gap. For Native EM $a>0$ and Serial EM $b$,
the overlap is bounded by
\begin{equation*}
\max\!\left(0,\frac{a+b-1}{a}\right)\le\mathrm{SCR}
\le\min\!\left(1,\frac{b}{a}\right).
\end{equation*}
Thus marginals bound the overlap, and paired observations identify it.
Also $|\mathrm{SCR}-b|\le1-a$: SCR equals Serial EM when Native is
perfect and is undefined when Native has no successes.

For a natural-task illustration, Qwen arithmetic FullFill and UnionFill
both solve 36/96 items but retain 35/94 and 36/94 Native successes.
This one-item difference illustrates distinct information, not a
controller advantage. A larger turnover appears in the named semantic
Qwen cell: 28 Native and 37 Serial successes overlap on six items,
so 22 losses and 31 gains produce a net gain of nine
(Table~\ref{tab:extended3}). Independent requests can have correlated
correctness through shared item difficulty. Copying an existing selection
is evaluated separately by plan fidelity. Table~\ref{tab:revision2}
compares models on \mbox{a common Native-correct population.}

\begin{algorithm}[t]
\caption{Paired Native and Serial evaluation.}
\label{alg:paired-evaluation}
\begin{algorithmic}[1]
\Require Problem \(x\), candidates \(A\), length \(L\), alphabet \(\Sigma\)
\Require Support \(s\in\{\textsc{Full},\textsc{Union},\textsc{Trie}\}\), fill flag \(f\)
\State \(n\gets\operatorname{Native}(x,A)\)
\State \(\hat y\gets\epsilon\)
\For{\(t=1,\ldots,L\)}
  \State \(V\gets\Sigma\)
  \If{\(s=\textsc{Union}\)}
    \State \(V\gets\{a_t:a\in A\}\)
  \ElsIf{\(s=\textsc{Trie}\)}
    \State \(V\gets\{a_t:a\in A,\ a_{<t}=\hat y\}\)
  \EndIf
  \If{\(V=\emptyset\)}
    \State \Return \(n,\textsc{Invalid}\)
  \EndIf
  \If{\(f\land |V|=1\)}
    \State \(z\gets\operatorname{sole}(V)\)
  \Else
    \State \(z\gets\operatorname{Local}(x,A,L,t,\hat y,V)\)
  \EndIf
  \If{\(z\notin V\)}
    \State \(z\gets\textsc{InvalidMarker}\)
  \EndIf
  \State \(\hat y\gets\hat y\mathbin{\|}z\)
\EndFor
\State \Return \(n,\hat y\)
\end{algorithmic}
\end{algorithm}

\subsection{Arithmetic and candidate
construction}\label{arithmetic-and-candidate-construction}

Positive operands have products of exactly L digits. Balanced operand lengths differ by at most one digit; unbalanced
lengths differ by at least two.
The grade school simulator records positive outgoing carries within each
multiplication row and the final partial product summation. Carry chain
length counts consecutive carry events within a row or summation,
resetting between them. Elementary multiplications, operand lengths,
separate multiplication/addition carries, full carry traces, and
zero/repeated digit counts are retained. Digit counts use concatenated operands.

For each length and structure, a frozen bounded pool defines low carry
at or below its median and high carry above it. Final selection takes 16
items from each structure/carry cell per length. Products are unique in
test. Pilot/development products and unordered operand pairs are
excluded. Simulator output is checked against integer multiplication.

Primary arithmetic candidates use a four by two construction matching an
accepted lexical set's vector of candidate counts compatible with the gold prefix. All have
equal length and no leading zero.

For prefix length k, K(k) counts compatible complete candidates. Stored
metadata includes K(0) through K(L), uniform candidate entropy log2
K(k), first unique prefix, mean entropy over k=0 through
L\ensuremath{-}1, and next character branch entropy. Every natural pair
has the same K vector along the gold path, with K(0)=8 and K(1)=2.
Branching outside the gold path and support at each position can differ.
Entropy quantifies uncertainty under \mbox{a uniform distribution over candidates.}

\subsection{Lexical sources and
review}\label{lexical-sources-and-review}

WordNet 3.0 supplies noun synsets and unchanged definitions. The wordfreq
3.1.1 supplies Zipf frequency. Targets are ASCII lowercase single words
of length three to eight with Zipf at least 2.5. Each source sense has
one eligible lemma. Screening excludes proper name senses, multiword
forms, target/form cues, shared synset candidates, adjacent taxonomic
senses and documented semantic conflicts. Records retain raw lemmas, same-length aliases, candidate senses and
possible outside answers. Acceptance applies to the specified sense within the closed candidate set.

Construction begins by sampling four groups by initial letter, with two
members each. All eight candidates require acceptable noun definition
sources and a frequency range at most 1.5 Zipf units. Gold is sampled
uniformly from the constructed set before duplicate rejection, review
and quota selection. Frequency strata split at Zipf 4.0. The presence of
repeated letters defines the second stratification factor. Frozen length
quotas select the first eligible bank ranks in each cell. Those
acceptance steps condition the final distribution. Controls using only
the candidate set audit remaining set cues. Development and test share no gold targets or unordered operand
pairs; distractors may recur.

Before formal experiments, two reviewers independently assessed all 576
bank items without prior answer records or model outputs. Both selected
the intended answer and judged it uniquely acceptable among the eight
candidates for all items. The planned resolution procedure was joint
source checking and discussion. No \mbox{answer/uniqueness adjudication was needed.}

Inclusion separately applies source/ambiguity rules: 543 items were
eligible, 33 held (26 for gloss clarity, seven for candidate overlap),
and 384 selected by frozen quotas. The remaining 159 eligible items were unused. For example, mug has an
underspecified gloss fragment, while granola competes with listed
oatmeal. Records \mbox{include full definitions and candidates.}

The independent calibration bank contains 2,400 previously unused
targets, all receiving LLM assisted lexical screening. Two humans
independently review 225 subjects: an audit of 192 presampled items, 32 per
length, plus 33 deduplicated targeted cases covering screening holds. In
the random audit, answer and uniqueness judgments agree throughout,
including one jointly nonunique item. Definition sufficiency differs on
15 items. Across both groups, 44 subjects receive a negative or
uncertain judgment from either reviewer. Their union with unresolved
screening holds excludes 53 subjects, leaving 2,347. Review records retain disagreements and identify all human-reviewed items.

Short eligible nouns are scarce after frequency, sense, target isolation
and candidate set requirements. Quotas based on available sources yield
61, 315, 442, 511, 548 and 470 retained subjects at lengths three
through eight. Quotas precede model outcomes, and every retained item is
run regardless of success. The main paired population remains balanced.
Common length standardization accounts for the cohorts' different
length mixtures.

\subsection{Output conditions and
stopping}\label{output-conditions-and-stopping}

Native makes one request with eight choices. Its control uses only the
candidates and withholds the problem. Independent and Serial query all L
positions with the same problem and candidates. Independent provides an
absolute position without a prefix. Serial supplies its actual generated
prefix. TeacherForced supplies the gold prefix. Gold Copy supplies the
target, and original SelfPlanSpell supplies that item's main Native
selection. Fidelity compares output with the selected plan. Gold
accuracy compares with the task answer.

The main content track uses ten digits or 26 lowercase letters and makes
no END request. The separate track with 51 choices uses 26 letters, ten
digits, eleven centrally configured punctuation symbols,
space/newline/tab and END on 72 preselected natural items. Punctuation
is defined by this project because the original demonstration did not enumerate
it. The cap is L+3 requests. Exact completion requires gold immediately
followed by END. Content EM and content in the first L positions are separately scored.
Exclusive stopping categories follow this priority: exact completion,
premature END, correct content without END, step cap exhaustion, wrong
content. Auxiliary indicators preserve overlapping exhaustion/content
errors.

The prefix intervention preselects one position per item and crosses
correct/error drafts with readout of the whole answer or a local character. An error
draft flips one preceding character. It need not remain compatible with
a complete candidate. Schedules retain the supplied states and are
independent of observed outcomes.

\subsection{Representation and readout
controls}\label{representation-and-readout-controls}

The isomorphic panel uses 20 bases per length, stratified by repetition
and early/late uniqueness. COPY/REVERSE and digit/letter encodings
create four presentations per base, with shared geometry and matched
order. The main bijection maps digits 0 through 9 to letters a through j. A second mapping to
etaoinshrd uses twelve fixed bases. The semantic panel has 128 source
facts, 16 per Boolean pattern, rendered as named fields or binary
encodings. Three two option decisions construct each output.
Field and binary representations use different wording and labels.

The single position panel has 120 natural parents and twelve symbolic
COPY bases with two encodings: 144 records, 132 independent clusters. It
supplies a gold string and preselected absolute position. Raw provides
string/position. Indexed adds the entire position table. Direct
additionally provides the extracted character. A true/false identity
check is separate. These representations change addressability, salience
and prompt length together.

Opaque, opaque remapped, readable aligned and readable conflicting names
retain the character descriptions. A permutation with no fixed points
rebinds names. Gold follows descriptions. Extra conflict loss is
(aligned\ensuremath{-}conflicting)\ensuremath{-}(opaque\ensuremath{-}remapped).
Jev has three repeats. Cross model estimates use the first two shared
repetition labels. This supplementary readout uses reported probability
argmax with displayed order tie breaking and retains returned
selections. StartLux's ordered JSON bypasses automatic renaming. Qwen
retains name/description mappings and records full vocabulary mass on
legal codes.

Matched ordinary Qwen generation crosses three evidence sources and
code/character outputs on all 144 records, giving 864 requests. Both
contracts use disabled thinking, greedy decoding, a budget of eight tokens,
identical evidence and matched order. Malformed outputs fail. Subsequent
marginal prompt/order checks and the full crossed grid retain the same
parents. Appendices~\ref{app:revision} and~\ref{app:prompts} give inference families
and verbatim contracts.

The fresh own plan panel selects five natural items per domain/length
without outcomes. Each decision model makes 60 fresh Native choices,
then copies its selected string under raw/indexed evidence and
Independent/Serial control. The indexed table is computed from that
selection, including errors. Each model uses 1,380 requests. Native is
shared across its four copying arms. Fidelity across all items, gold accuracy,
SCR conditioned on fresh Native correctness, errors, survival and costs use these
plans, separately from original SelfPlanSpell. Four indexing contrasts and four controller--representation
interactions form separate families for Holm correction.

\subsection{Confidence and statistical
inference}\label{confidence-and-statistical-inference}

Chosen path product multiplies reported probabilities of emitted
characters. Geometric mean exponentiates the mean log probability.
Minimum top choice probability takes the smallest per step maximum and
can differ from chosen path scores when returned choices disagree with
argmax. TeacherForced also records gold path product and mean. Brier,
ten fixed equal width reliability bins/ECE, AUROC and average precision
evaluate completed string correctness. One class discrimination is
undefined. Evaluation preserves original scores and exact floating-point
ties across cohorts.

Native to position projection sums complete candidate probabilities by
character and compares them with independent queries. Raw mass and
separately normalized TV/JS are labelled. Unsupported mass covers
characters absent from every candidate at that position. Its uniform
reference is one minus support size divided by alphabet size. Bootstrap uses item-averaged position scores.

Original inference uses 99,999 bootstrap draws, seed 20261005. Natural
strata are domain/length. Symbolic bases use length, repetition and
profile. Semantic bases use gold pattern. Conditional denominators are
recomputed, with empty ones undefined. Boundary counts receive
supplementary conditional binomial intervals. For the zero success
Qwen3-8B cells in Table~\ref{tab:primary}, the two sided exact upper
bounds are 6.1\% for 0/59 and 1.0\% for 0/363. Empirical bootstrap intervals with zero width describe the observed sample boundary. The primary six
Native/Serial exact McNemar tests share a Holm correction at 0.05. Later
panels state their own seeds, draw counts and correction families.
Character observations remain grouped by item.
Exploratory GEE fits cluster by item \citep{liang1986longitudinal}. Failed, quasi separated or nonfinite fits are not used for inference.

\subsection{Models and
reproducibility}\label{models-and-reproducibility}

Jev is pinned to jev-1.13.0 with SDK 0.7.0. Formal construction,
independent calibration and candidate support studies score the
choice returned by the service. The readout/own plan diagnostics predefine
probability argmax. Both are retained with reported rounding and
disagreements. Separate repeat records capture hosted response variation.

StartLux-Decision-4B uses BF16 revision
\nolinkurl{9302aedb7f7bd889994336f2ae919dffbf5dbee1} and code
\nolinkurl{c61bfab3bb4fa3ae9d155d457247a7fd0a299a24}, Torch 2.8/CUDA
12.8 and Transformers 5.8.1. Released temperature, wide choice grouping
and the slow Torch fallback are preserved. Qwen3-8B uses full BF16 (checkpoint digest
\nolinkurl{d45786379a66d7e29f82e4f694b5eab47fe7c15ce85777b3b8b0fa30fb73e7e2}),
disabled thinking and Transformers 4.57.1 with the same Torch/CUDA
version. Legal code tokens are checked at the actual prompt boundary.
Qwen3.5-27B uses BF16 revision
\nolinkurl{fc05daec18b0a78c049392ed2e771dde82bdf654},
Torch 2.8/CUDA 12.8 and Transformers 5.17.0. Direct queries allow eight
new tokens per character, 32 per string and 256 per dialogue JSON.
Thinking is disabled. Only outer whitespace is stripped; extra text,
wrong case, illegal values and truncation fail. Checkpoint, tokenizer
and source hashes accompany the responses.

Frozen schedules interleave domains and conditions within collection
windows. Logs retain exact prompts, candidate/option orders, prefixes,
references, selections, available probabilities, service confidence,
model/SDK identity, timing, errors and retries. Generated outputs
without probability maps are explicitly marked.

Costs report logical requests, input tokens, available physical forwards
and summed recorded request times within each recorded runtime. Timing
covers requests and excludes loading and orchestration overhead. Hosted
and local runs use different hardware. Jev costs are calculated from
recorded usage at the frozen rate of USD 0.042 per million input tokens.
The reported values represent estimated charges under this fixed rate.
Own-plan costs comprise one Native and L copy calls; experimental
spend shares Native across arms.

The reproducibility materials contain frozen items, schedules, complete
requests/responses, lexical review records and offline reconstruction.
The evidence index maps exhibits to data and analyses; verbatim
prompts and contracts appear in Appendix~\ref{app:prompts}.

\FloatBarrier
\setcounter{table}{0}
\renewcommand{\thetable}{B\arabic{table}}
\section{Character Controls}\label{app:results}\label{controlled-character-construction}
\begin{figure*}[tbp]
\centering
\includegraphics[width=.86\textwidth]{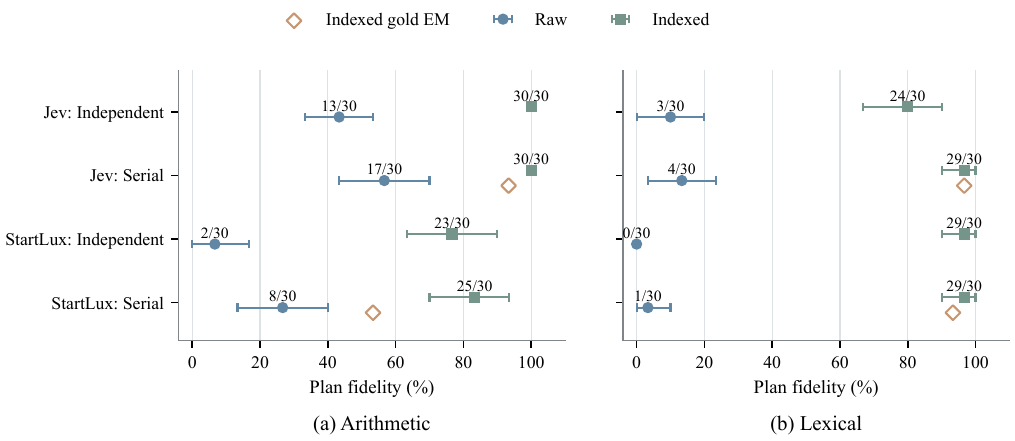}
\caption{Indexing improves fidelity to selected plans without guaranteeing task correctness. Each panel uses 30 fresh plans per domain/model. (a) Arithmetic fidelity can exceed gold EM because incorrect plans are copied faithfully; (b) lexical fidelity also improves for both models. Intervals are 95\% item bootstrap.}
\label{fig:ownplan}
\end{figure*}

\subsection{Prefix feedback and supplied targets}
\label{prefix-feedback-and-supplied-targets}

Independent omits the draft; Serial supplies its generated prefix;
TeacherForced supplies the gold prefix on 60 preselected items per domain.
Character accuracy is averaged within item. TeacherForced string success
requires every gold-prefix-conditioned decision to be correct.

Fresh-plan copying uses 30 Native selections per model/domain, shared
across Raw/Indexed and Independent/Serial.
The position table is computed from the selected string, including
errors. Fidelity scores that string; gold EM scores the original task
(Figure~\ref{fig:ownplan}). Holm correction treats four indexing contrasts and four
controller--representation interactions as separate families.

\subsection{Three workflows on the same selected-plan parents}
\label{same-parent-choice-workflows}

On the original 60 selected-plan parents per domain/model, program
return emits the recorded Native choice with perfect fidelity and no
additional call. Independent Serial solves without receiving that
choice; SelfPlanSpell copies it through $L$ character queries.
Jev arithmetic copying is faithful on 40/60 but gold-correct on 38/60:
two incorrect selections are faithfully preserved. The workflows share parents, with Native and Serial independently
solving each primary item.

\subsection{Natural whole-answer generation}
\label{natural-whole-answer-generation}

Qwen3-8B receives each of the 384 original problems and candidate sets
and emits the complete answer in one greedy response, with thinking
disabled and 64 new tokens. No selected or gold answer is supplied.
The same checkpoint and items are used in Native and legal-code Serial;
the response contract changes to ordinary text. Joint, lost and gained
counts below retain this pairing.

\begin{table}[tbp]
\centering\small
\caption{Whole-answer generation improves Qwen3-8B accuracy while changing which items succeed (384 per domain). Serial legal-code EM is zero in both domains. Joint, lost and gained counts compare independent Native and whole-generation outcomes.}
\label{tab:natural-whole-generation}
\setlength{\tabcolsep}{3pt}
\begin{tabular}{@{}lrrrrr@{}}
\toprule
Domain & Native & Whole & Joint & Lost & Gained\\
\midrule
Arithmetic & 59 & 181 & 41 & 18 & 140\\
Lexical & 363 & 372 & 358 & 5 & 14\\
\bottomrule
\end{tabular}
\end{table}

\subsection{Digit and letter representations and semantic controls}
\label{digitletter-and-semantic-controls}

COPY/REVERSE uses 120 shared source bases. Mapping digits 0--9 to
letters a--j preserves strings, candidate geometry and ten local choices;
both operations are averaged within base for inference. The semantic
panel uses 128 shared fact records, 16 per Boolean pattern, rendered as
named fields or binary strings. Each local output requires three
binary decisions. SCR conditions on representation-specific Native successes;
all-item EM scores the full population of 128 fact records.

\begin{figure*}[tbp]
\centering
\includegraphics[width=.98\textwidth]{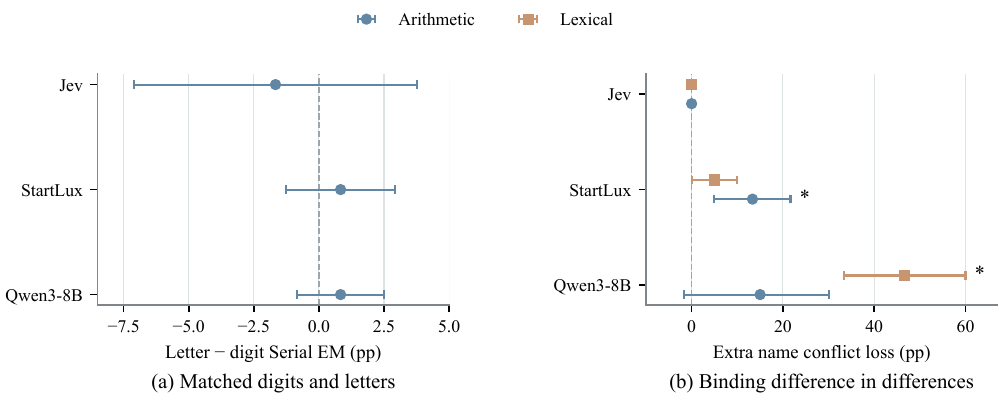}
\caption{Matched symbol operations yield small letter--digit contrasts, while option-name conflicts remain interface dependent. (a) On 120 shared COPY/REVERSE bases, contrast intervals span both directions. (b) Extra conflict losses survive correction for StartLux arithmetic and Qwen3-8B lexical readout. Paired intervals are 95\%; asterisks mark Holm $p<0.05$ after correction within each comparison family.}
\label{fig:representation-controls}\label{tab:extended2}\label{tab:extended5}
\end{figure*}

\begin{table*}[tbp]
\centering
\small
\caption{Jev preserves most correct semantic decisions under either encoding. SCR reports retained/correct Native counts and 95\% intervals separately for named and binary outputs.}
\label{tab:extended3}
\setlength{\tabcolsep}{3pt}
\begin{tabular}{@{}llrrrr@{}}
\toprule
Model & Representation & Native EM & Serial EM & SCR count & SCR {[}95\% CI{]} \\
\midrule
Jev & named & 89.8 & 90.6 & 111/115 & 96.5 {[}93.0, 99.2{]} \\
Jev & binary & 68.0 & 85.2 & 80/87 & 92.0 {[}86.0, 96.8{]} \\
StartLux & named & 51.6 & 71.1 & 51/66 & 77.3 {[}67.6, 86.6{]} \\
StartLux & binary & 39.1 & 29.7 & 25/50 & 50.0 {[}40.4, 60.5{]} \\
Qwen3-8B & named & 21.9 & 28.9 & 6/28 & 21.4 {[}8.0, 36.0{]} \\
Qwen3-8B & binary & 14.8 & 15.6 & 1/19 & 5.3 {[}0.0, 16.7{]} \\
\bottomrule
\end{tabular}
\end{table*}

\subsection{Readout at a single position and option binding}
\label{single-position-readout-and-option-binding}

Sixty parents per domain and twelve symbolic bases each supply a target
and one preselected position. Raw provides string and position; Indexed
adds the position table; Direct also supplies the extracted character.
Cross-model estimates average two shared repeats within parent; Jev's
three-repeat estimate is separate. Qwen scores legal codes by conditional
softmax. Opaque remapping, readable alignment and readable conflict
preserve character criteria. Extra conflict loss is
$(\mathrm{aligned}-\mathrm{conflicting})-(\mathrm{opaque}-\mathrm{remapped})$.
Figure contrasts and cross-model contrasts \mbox{receive separate Holm corrections.}

\begin{figure*}[tbp]
\centering
\includegraphics[width=.98\textwidth]{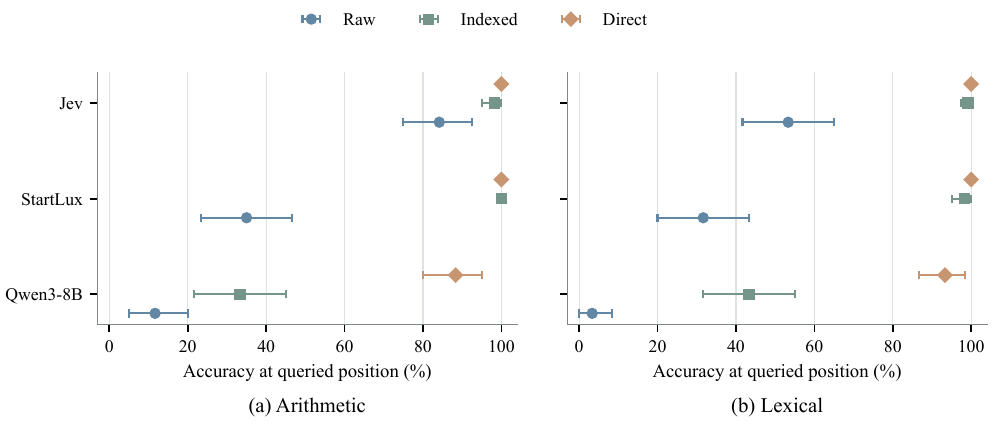}
\caption{On 60 supplied-target parents per domain, (a,b) Jev and StartLux approach perfect indexed readout from lower raw accuracy; Qwen3-8B legal-code accuracy stays lower. Intervals are 95\% item bootstrap.}
\label{fig:readout-evidence}\label{tab:extended4}
\end{figure*}

\begin{figure*}[tbp]
\centering
\includegraphics[width=.98\textwidth]{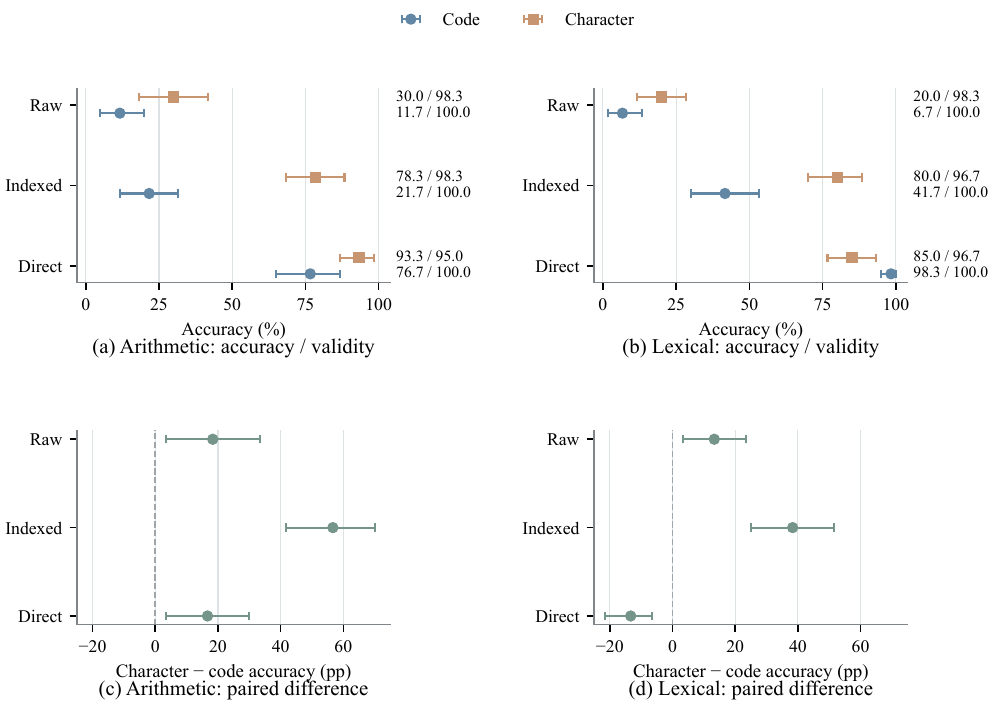}
\caption{The better output contract depends on evidence presentation. Under matched greedy Qwen3-8B requests, (a,b) format validity stays high despite accuracy differences. (c,d) Indexing favors characters in both domains, whereas direct lexical evidence favors codes. Malformed responses fail; paired intervals are 95\%.}
\label{fig:matched-generation}\label{tab:extended6}
\end{figure*}

\subsection{Matching generation contracts}
\label{matching-generation-contracts}

Code and character requests use identical evidence and order, disabled
thinking, greedy decoding and eight new tokens. Malformed responses fail.
The 144-record panel crosses three evidence forms with two contracts;
all repetitions stay within parent. The full prompt/order grid is in
Appendix~\ref{crossed-prompt-and-option-order-robustness}; exact contracts
appear in Appendix~\ref{app:prompts}.

\subsection{Distribution, stopping and cost}
\label{distribution-stopping-and-cost}

Candidate probabilities are projected onto each position by summing
complete-answer mass for each character. TV/JS use separately normalized
distributions; raw mass is retained. Unsupported mass is compared with
$1-|\mathrm{support}|/|\Sigma|$, and positions are averaged within item.

The separate 51-choice track uses 72 preselected parents and a cap of
$L+3$ requests. It scores correct END, premature END, correct content
without END, cap exhaustion and wrong content under the priority in
Appendix~\ref{output-conditions-and-stopping}. Request times exclude
loading and orchestration; costs count logical requests and recorded
usage. Native followed by program return needs one request.

\setcounter{table}{0}
\renewcommand{\thetable}{C\arabic{table}}
\section{Dialogue Evaluation}\label{app:extensions}
\label{larger-model-controls-and-prior-state-replay}
\subsection{Prior-state access}\label{prior-state-access-in-dialogue-replay}
Schema-Guided Dialogue \citep{rastogi2020towards} supplies simulated
English dialogues with crowdsourced paraphrases (CC BY-SA 4.0).
The 96 test prefixes contain four, six or eight user turns (32 each),
cover ten services (at most twelve dialogues per service), and have one
to three categorical slots plus active intent. A service persists
throughout each prefix; selection requires state changes and a nonempty
categorical value. Twelve development prefixes are disjoint by split,
identifier and content.

HistoryRecompute receives the transcript and schema. Rollout receives
its previous predicted state and the latest SYSTEM/USER exchange;
Oracle substitutes the gold prior state. Initial intent is NONE,
absent slots UNSET and explicit indifference DONTCARE. Current gold
state is reserved for scoring. Jev returns multiple typed fields,
StartLux retains semantic names/descriptions, and Qwen emits strict
JSON with precisely the requested keys and legal values.

Mean-turn EM weights dialogues equally. Trajectory and final EM score
complete prefixes and terminal states. Each model has 1,924 logical
requests including perturbations and 1,800 physical responses; identical
wire inputs share 124 rows. Six rows connect separate dialogues into
94 response-sharing components. Bootstrap either resamples complete
dialogues or these components, retaining \mbox{the scores for receiving dialogues.}

\begin{figure}[t]
\centering
\includegraphics[width=\columnwidth]{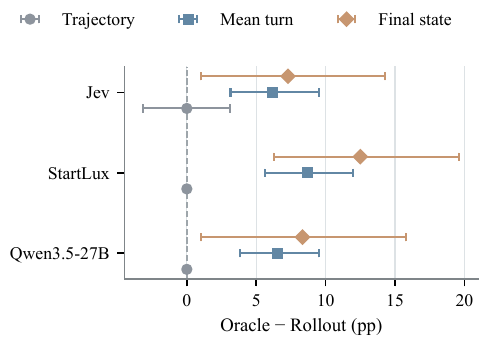}
\caption{Identical trajectory accuracy conceals differences in later states. On 96 dialogues per model, gold prior states improve mean-turn and final-state accuracy over predicted prior states, while trajectory EM is unchanged. Exploratory 95\% intervals resample dialogue groups linked by shared responses.}
\label{fig:prior-state}
\end{figure}

\begin{table*}[tbp]
\centering
\small
\caption{Equal trajectory accuracy coexists with different later-state accuracy (96 dialogues per arm). Mean turn EM weights dialogues equally. Trajectory and final EM evaluate complete trajectories and terminal states.}
\label{tab:extension5}
\setlength{\tabcolsep}{3pt}
\begin{tabular}{@{}llrrrr@{}}
\toprule
Model & Mode & Trajectory & Mean turn & Final & Valid format \\
\midrule
Jev & HistoryRecompute & 27.1 & 75.3 & 53.1 & 100.0 \\
Jev & Oracle & 33.3 & 83.2 & 68.8 & 100.0 \\
Jev & Rollout & 33.3 & 77.0 & 61.5 & 100.0 \\
StartLux & HistoryRecompute & 25.0 & 69.7 & 59.4 & 100.0 \\
StartLux & Oracle & 30.2 & 80.6 & 71.9 & 100.0 \\
StartLux & Rollout & 30.2 & 71.9 & 59.4 & 100.0 \\
Qwen3.5-27B & HistoryRecompute & 30.2 & 76.3 & 58.3 & 100.0 \\
Qwen3.5-27B & Oracle & 26.0 & 80.0 & 61.5 & 100.0 \\
Qwen3.5-27B & Rollout & 26.0 & 73.5 & 53.1 & 100.0 \\
\bottomrule
\end{tabular}
\end{table*}

\begin{table*}[tbp]
\centering
\small
\caption{Gold prior states improve mean-turn and final-state accuracy without changing trajectory EM. Differences are Oracle minus Rollout (percentage points), with paired-dialogue and response-component 95\% intervals.}
\label{tab:extension6}
\setlength{\tabcolsep}{3pt}
\begin{tabular}{@{}llrr@{}}
\toprule
Model & Endpoint & Difference {[}95\% CI{]} & Component 95\% CI \\
\midrule
Jev & trajectory & 0.0 {[}-2.1, 2.1{]} & {[}-3.1, 3.1{]} \\
Jev & mean turn & 6.2 {[}3.7, 8.6{]} & {[}3.1, 9.5{]} \\
Jev & final & 7.3 {[}2.1, 12.5{]} & {[}1.0, 14.3{]} \\
Jev & mean field & 2.4 {[}1.5, 3.3{]} & {[}1.3, 3.6{]} \\
StartLux & trajectory & 0.0 {[}0.0, 0.0{]} & {[}0.0, 0.0{]} \\
StartLux & mean turn & 8.7 {[}6.7, 10.7{]} & {[}5.6, 12.0{]} \\
StartLux & final & 12.5 {[}7.3, 17.7{]} & {[}6.2, 19.6{]} \\
StartLux & mean field & 3.7 {[}2.9, 4.6{]} & {[}2.4, 5.2{]} \\
Qwen3.5-27B & trajectory & 0.0 {[}0.0, 0.0{]} & {[}0.0, 0.0{]} \\
Qwen3.5-27B & mean turn & 6.5 {[}4.3, 8.9{]} & {[}3.9, 9.5{]} \\
Qwen3.5-27B & final & 8.3 {[}2.1, 14.6{]} & {[}1.0, 15.8{]} \\
Qwen3.5-27B & mean field & 3.0 {[}2.0, 4.1{]} & {[}1.7, 4.5{]} \\
\bottomrule
\end{tabular}
\end{table*}

\begin{table*}[tbp]
\centering
\small
\caption{Failed trajectories can recover correct later or final states. Error populations vary by model and arm.}
\label{tab:extension7}
\setlength{\tabcolsep}{3pt}
\begin{tabular}{@{}llrrr@{}}
\toprule
Model & Mode & Failed & Any recovery & Final recovery \\
\midrule
Jev & HistoryRecompute & 70 & 36 & 25 \\
Jev & Oracle & 64 & 47 & 34 \\
Jev & Rollout & 64 & 36 & 27 \\
Qwen3.5-27B & HistoryRecompute & 67 & 37 & 27 \\
Qwen3.5-27B & Oracle & 71 & 47 & 34 \\
Qwen3.5-27B & Rollout & 71 & 33 & 26 \\
StartLux & HistoryRecompute & 72 & 47 & 33 \\
StartLux & Oracle & 67 & 56 & 40 \\
StartLux & Rollout & 67 & 38 & 28 \\
\bottomrule
\end{tabular}
\end{table*}

\begin{figure}[t]
\centering
\includegraphics[width=\columnwidth]{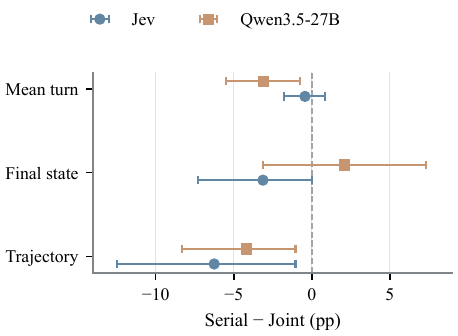}
\caption{The accuracy effect of local field construction differs across models. On 96 dialogues with identical visible history and schemas, Qwen3.5-27B shows a corrected mean-turn loss, while Jev's interval spans zero. Final-state intervals span zero for both models. Intervals are 95\% dialogue bootstrap.}
\label{fig:workflow}
\end{figure}

\subsection{Matched current-state construction}
\label{matched-construction-of-categorical-dialogue-states}
Joint, Independent and Serial use the same visible history and schema
on 96 dialogues, 576 user turns and 1,716 field turns. Joint requests all
fields together; Independent requests each separately; Serial also
receives earlier field predictions for that turn. Neither local arm
receives Joint's output. Field order uses seed 20261009.

All arms allow 256 new tokens, disabled thinking and greedy decoding;
invalid JSON, extra keys, illegal values and truncation fail. The primary
endpoint averages joint state turn EM within dialogue, then across
dialogues. Two Serial-minus-Joint sign tests share Holm correction
(99,999 draws, plus-one correction). State retention is dialogue-weighted
local/Joint jointly-correct mass divided by Joint-correct mass, distinct
from character SCR. Conditional denominators are redrawn. Final-state and trajectory intervals are descriptive. Each request
receives a separate response.

\begin{table*}[tbp]
\centering
\small
\caption{Local field construction reduces Qwen mean-turn accuracy; the Jev contrast spans zero. Accuracy (\%) has 95\% dialogue bootstrap intervals. Dialogues receive equal mean-turn weight. Final-state EM scores the terminal dialogue state; trajectory EM requires correctness at every turn.}
\label{tab:revision25}
\setlength{\tabcolsep}{3pt}
\begin{tabular}{@{}llrrr@{}}
\toprule
Model & Arm & Mean turn & Final state & Trajectory \\
\midrule
Jev & Joint & 76.0 {[}71.0, 80.7{]} & 55.2 {[}44.8, 64.6{]} & 33.3 {[}24.0, 42.7{]} \\
Jev & Independent & 76.0 {[}71.0, 80.8{]} & 56.2 {[}45.8, 65.6{]} & 32.3 {[}22.9, 41.7{]} \\
Jev & Serial & 75.5 {[}70.8, 80.0{]} & 52.1 {[}41.7, 62.5{]} & 27.1 {[}18.8, 36.5{]} \\
Qwen3.5-27B & Joint & 75.4 {[}70.8, 79.8{]} & 57.3 {[}46.9, 66.7{]} & 28.1 {[}19.8, 37.5{]} \\
Qwen3.5-27B & Independent & 70.6 {[}66.1, 75.0{]} & 58.3 {[}47.9, 67.7{]} & 20.8 {[}12.5, 29.2{]} \\
Qwen3.5-27B & Serial & 72.3 {[}67.9, 76.6{]} & 59.4 {[}49.0, 68.8{]} & 24.0 {[}15.6, 32.3{]} \\
\bottomrule
\end{tabular}
\end{table*}

\begin{table*}[tbp]
\centering
\small
\caption{High state retention still requires more local requests. Format validity is in percent. Retention intervals weight dialogues equally and include raw retained turn counts.}
\label{tab:revision27}
\setlength{\tabcolsep}{3pt}
\begin{tabular}{@{}llrrrrr@{}}
\toprule
Model & Arm & Retained turns & Retention {[}95\% CI{]} & Valid & Calls & Input tokens \\
\midrule
Jev & Independent & 423/429 & 98.5 {[}97.2, 99.6{]} & 100.0 & 1716 & 1651282 \\
Jev & Joint & reference & reference & 100.0 & 576 & 822825 \\
Jev & Serial & 420/429 & 97.9 {[}96.6, 99.2{]} & 100.0 & 1716 & 1686253 \\
Qwen3.5-27B & Independent & 388/426 & 91.9 {[}88.9, 94.7{]} & 99.2 & 1716 & 1059954 \\
Qwen3.5-27B & Joint & reference & reference & 100.0 & 576 & 591013 \\
Qwen3.5-27B & Serial & 395/426 & 93.3 {[}90.6, 95.8{]} & 99.4 & 1716 & 1084404 \\
\bottomrule
\end{tabular}
\end{table*}

\paragraph{Perturbation schedule.}\label{dialogue-perturbations}
Twelve test dialogues receive one task-preserving paraphrase; six of
these also receive one permutation of displayed semantic values.
Comparisons preserve dialogue clusters and shared responses across arms.

\setcounter{table}{0}
\renewcommand{\thetable}{D\arabic{table}}
\section{Supporting Results and Robustness}\label{app:revision}\label{additional-diagnostics-and-robustness}

\subsection{Shared Native-correct population}
\label{full-population-reconstruction}

The primary population contains 384 items per domain and 64 per
length for each model. The shared-correct comparison below conditions
on items both Jev and direct-output Qwen select correctly. Intervals use 99,999 length-stratified paired item bootstrap draws
with seed 20261008.

\begin{table*}[tbp]
\centering
\small
\caption{The arithmetic--lexical contrast persists on a common Native-correct population. Coverage, retained counts and 95\% item bootstrap intervals stratified by length are shown.}
\label{tab:revision2}
\setlength{\tabcolsep}{3pt}
\begin{tabular}{@{}llrrr@{}}
\toprule
Model & Domain & Coverage & Retained & SCR {[}95\% CI{]} \\
\midrule
Jev & math & 367/384 & 118/367 & 32.2 {[}28.1, 36.2{]} \\
Qwen3.5-27B & math & 367/384 & 38/367 & 10.4 {[}7.4, 13.4{]} \\
Jev & lexical & 379/384 & 8/379 & 2.1 {[}0.8, 3.7{]} \\
Qwen3.5-27B & lexical & 379/384 & 131/379 & 34.6 {[}29.8, 39.3{]} \\
\bottomrule
\end{tabular}
\end{table*}

\subsection{Repeated Native observations}
\label{repeated-native-observations}

Seventy-two parents (six per domain/length) receive three identical
Native requests and three Serial trajectories under pinned Jev.
Repeats remain grouped by parent. Choice stability and probability stability have distinct endpoints.

\begin{table*}[tbp]
\centering
\small
\caption{Across three repeats, Native choices agree but Serial successes vary. Counts are retained/Native correct.}
\label{tab:revision3}
\setlength{\tabcolsep}{3pt}
\begin{tabular}{@{}llrrr@{}}
\toprule
Domain & Native correct & Serial 0 & Serial 1 & Serial 2 \\
\midrule
math & 35/36 & 12/35 & 13/35 & 12/35 \\
lexical & 36/36 & 0/36 & 1/36 & 1/36 \\
\bottomrule
\end{tabular}
\end{table*}

\subsection{Natural task robustness under complete trajectories}
\label{natural-task-robustness-under-complete-trajectories}

The 72-parent natural panel varies two paraphrases, two option orders
and two additional trajectories separately from the reference, yielding
seven settings per model. Each setting has its own Native conditioning;
fixed-reference conditioning is additionally retained. Complete
trajectories use actual prefixes and whole-item paired intervals.
These natural solving requests receive no target answer, unlike
the supplied-target readout panel.

\begin{figure*}[tbp]
\centering
\includegraphics[width=.98\textwidth]{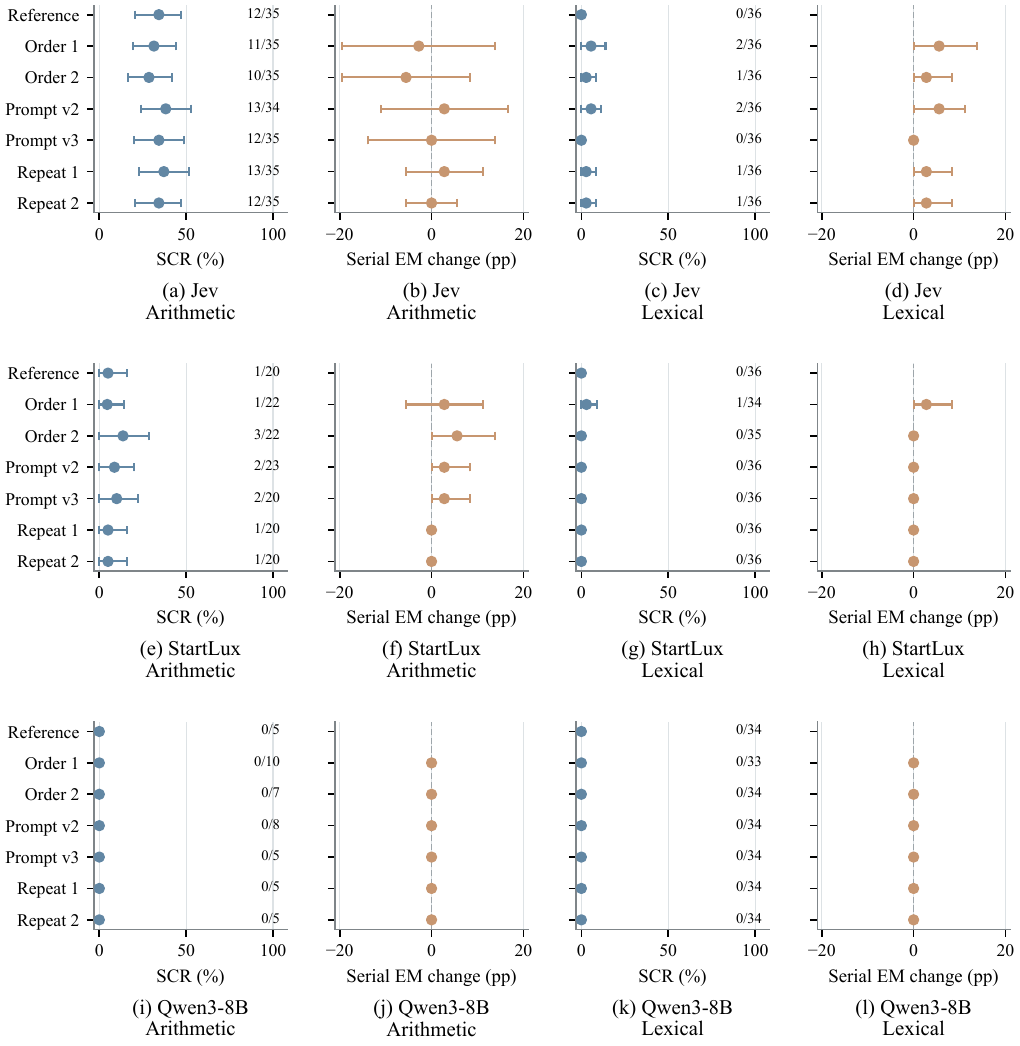}
\caption{The Jev domain gap survives natural-task perturbations. Across 72 items, (a,c) Jev retains more arithmetic than lexical answers; (e,g) StartLux retention stays low; (i,k) Qwen3-8B legal-code construction has no successes. Paired changes are imprecise. SCR uses setting-specific Native successes, with 95\% item bootstrap intervals.}
\label{fig:natural-robustness}
\end{figure*}

\subsection{Direct output natural prompt robustness}
\label{direct-output-natural-prompt-robustness}

Qwen3.5-27B uses 96 parents per domain, 16 per length selected before
outcomes, under three interleaved templates. Each has its own Native
request and Full Serial trajectory. Bootstrap retains templates within
parent/length; four nonreference Serial contrasts share one exact
McNemar/Holm family. All responses satisfy the character format.
The verbatim instructions appear in Appendix~\ref{app:prompts}.

\begin{table*}[tbp]
\centering
\small
\caption{Prompt wording changes Qwen content accuracy under valid character output. SCR uses template-specific Native successes. Changes from v1 have paired 95\% intervals and four-test Holm correction.}
\label{tab:revision28}
\setlength{\tabcolsep}{3pt}
\begin{tabular}{@{}llrrrrr@{}}
\toprule
Domain & Prompt & Serial EM & Retained & SCR {[}95\% CI{]} & EM change {[}95\% CI{]} & Holm p \\
\midrule
lexical & v1 & 39.6 & 38/95 & 40.0 {[}30.5, 49.5{]} & reference & N/A \\
lexical & v2 & 56.2 & 54/96 & 56.2 {[}46.9, 65.6{]} & 16.7 {[}7.3, 26.0{]} & 0.005 \\
lexical & v3 & 50.0 & 47/95 & 49.5 {[}39.6, 59.4{]} & 10.4 {[}1.0, 19.8{]} & 0.105 \\
math & v1 & 7.3 & 7/95 & 7.4 {[}3.1, 12.5{]} & reference & N/A \\
math & v2 & 28.1 & 27/94 & 28.7 {[}21.5, 35.8{]} & 20.8 {[}13.5, 28.1{]} & \textless0.001 \\
math & v3 & 13.5 & 12/95 & 12.6 {[}7.3, 18.8{]} & 6.2 {[}-1.0, 13.5{]} & 0.180 \\
\bottomrule
\end{tabular}
\end{table*}

\label{observed-prompt-setting-summaries}

Means average the seven tested marginal settings in the common
72-parent panel or the three templates in the separate Qwen panel.
They are not expectations over an unspecified prompt distribution.
Worst-setting SCR and conditional denominators are recomputed in each
parent bootstrap draw. Within each original setting, SCR and Serial EM
agree on model order: Jev leads all seven arithmetic settings and
five lexical settings; two lexical settings have no successes.

\begin{table}[H]
\centering\small
\caption{Worst-template retention remains well below Native accuracy. Direct Qwen3.5-27B SCR (\%) has 95\% whole-item bootstrap intervals. Each domain resamples 96 parents and recomputes its worst setting.}
\label{tab:prompt-aggregate}
\setlength{\tabcolsep}{3pt}
\begin{tabular}{@{}lll@{}}
\toprule
Domain & Mean [95\% CI] & Worst [95\% CI]\\
\midrule
Arithmetic & 16.2 [12.1, 20.6] & 7.4 [3.1, 11.7]\\
Lexical & 48.6 [40.3, 56.9] & 40.0 [30.5, 49.5]\\
\bottomrule
\end{tabular}
\end{table}

\subsection{Candidate supported local construction}
\label{candidate-supported-local-construction}

All 768 natural parents receive fresh Native, Full Serial and Union
Serial requests. Union filters the seeded alphabet to characters that
occur at the current position in any displayed candidate, preserving
choice identities and order. Singleton unions are filled with no call
or assigned probability. Both local arms use their own actual prefix;
neither queries END. Domains and conditions are interleaved.
Jev scores returned choices, StartLux preserves semantic options and
Qwen emits characters directly. Six Full/Union McNemar tests share Holm correction. Intervals use
99,999 length-stratified bootstrap draws with seed 20261009.

The tables distinguish paired recoveries/losses, actual requests,
candidate membership and Cartesian support. A position union can leave
the complete candidate set; its constraint differs from Trie.
Only two arithmetic positions have non-singleton unions; every
lexical position does.

\begin{figure*}[tbp]
\centering
\includegraphics[width=.97\textwidth]{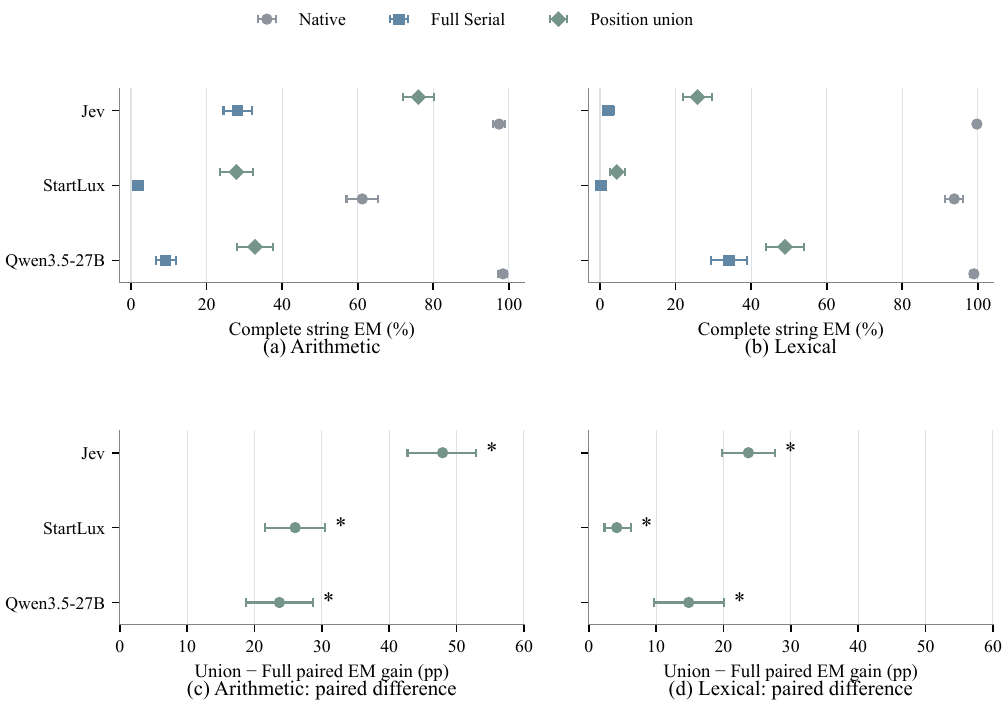}
\caption{Candidate support improves Serial success while leaving a gap to Native. On 384 formal items per domain, (a,c) arithmetic gains combine option restriction with deterministic filling; (b,d) lexical gains retain a model request at every position. Intervals are paired 95\% item bootstrap; asterisks indicate Holm $p<0.05$ across six tests.}
\label{fig:candidate-support}\label{tab:revision21}
\end{figure*}

\begin{table*}[tbp]
\centering
\small
\caption{Union recovers failed items while losing some Full successes (384 items per domain). Recovered means Full wrong/Union correct; lost means Full correct/Union wrong. Counts expose changes hidden by net accuracy.}
\label{tab:support-transitions}
\setlength{\tabcolsep}{5pt}
\begin{tabular}{@{}llrrrr@{}}
\toprule
Model & Domain & Both wrong & Recovered & Lost & Both correct \\
\midrule
Jev & math & 84 & 192 & 8 & 100 \\
Jev & lexical & 285 & 91 & 0 & 8 \\
StartLux & math & 274 & 103 & 3 & 4 \\
StartLux & lexical & 366 & 17 & 1 & 0 \\
Qwen3.5-27B & math & 244 & 105 & 14 & 21 \\
Qwen3.5-27B & lexical & 166 & 87 & 30 & 101 \\
\bottomrule
\end{tabular}
\end{table*}

\begin{table*}[tbp]
\centering
\small
\caption{Position unions reduce arithmetic calls while permitting noncandidate lexical strings. Membership percentages and recorded model request counts exclude retries.}
\label{tab:revision22}
\setlength{\tabcolsep}{3pt}
\begin{tabular}{@{}llrrrrr@{}}
\toprule
Model & Domain & Full calls & Union calls & Forced & Full member & Union member \\
\midrule
Jev & lexical & 2112 & 2112 & 0 & 2.6 & 28.6 \\
Jev & math & 2112 & 768 & 1344 & 33.9 & 100.0 \\
StartLux & lexical & 2112 & 2112 & 0 & 1.0 & 9.1 \\
StartLux & math & 2112 & 768 & 1344 & 5.2 & 100.0 \\
Qwen3.5-27B & lexical & 2112 & 2112 & 0 & 40.6 & 62.8 \\
Qwen3.5-27B & math & 2112 & 768 & 1344 & 20.1 & 98.2 \\
\bottomrule
\end{tabular}
\end{table*}

\begin{table*}[tbp]
\centering
\small
\caption{Matching gold-prefix ambiguity leaves much larger lexical Cartesian output spaces. Mean U averages union sizes within items; Forced counts deterministic positions; Cartesian support multiplies position-wise sizes.}
\label{tab:revision23}
\setlength{\tabcolsep}{3pt}
\begin{tabular}{@{}llrrrrrr@{}}
\toprule
Domain & Length & Items & Mean U & Forced & Min size & Median & Max \\
\midrule
math & 3 to 8 & 384 & 1+4/L & L\ensuremath{-}2 & 8 & 8 & 8 \\
lexical & 3 & 64 & 5.22 & 0.0 & 64 & 134 & 224 \\
lexical & 4 & 64 & 5.43 & 0.0 & 336 & 812 & 1,568 \\
lexical & 5 & 64 & 5.73 & 0.0 & 1,800 & 5,184 & 12,544 \\
lexical & 6 & 64 & 5.60 & 0.0 & 9,600 & 26,040 & 57,624 \\
lexical & 7 & 64 & 5.68 & 0.0 & 31,104 & 176,400 & 614,656 \\
lexical & 8 & 64 & 5.80 & 0.0 & 201,600 & 1,128,960 & 2,370,816 \\
\bottomrule
\end{tabular}
\end{table*}

\subsection{Matching complete feasible outputs and controller calls}
\label{matched-feasible-output-control}

Jev evaluates all 768 parents; Qwen uses 192 selected by item hash
before outcomes (16 per domain/length), with every arm of an item on
one instance. Seed 20261010 interleaves conditions within position rounds.
Full uses the domain alphabet. UnionCall uses position unions;
TrieCall restricts choices to continuations of candidates compatible
with the generated prefix, preserving exactly eight feasible outputs.
Call queries singletons. UnionFill/TrieFill instead supply them;
FullFill supplies position-union singletons and otherwise uses Full.
A wrong Trie branch can exclude gold. Malformed Trie prefixes abort,
and later positions are logged with zero calls. Membership is checked for valid completions. Confidence is recorded
for model calls, not deterministic fills.

Instructions and filtered option order are fixed. Eight TrieCall versus
Full/Union contrasts share one Holm family; twelve filling contrasts
form a separate family. Bootstrap uses 9,999 whole-item draws within
length. The exploratory domain contrast pairs the 384 gold-prefix-
geometry partners. Jev's lexical-minus-arithmetic EM is $-26.3$ points
[$-30.2$,$-22.4$] under Full and 2.6 [$-2.9$,7.8] under TrieCall.
The TrieCall interval includes both a negative and a positive
domain contrast.

\begin{figure*}[!t]
\centering
\includegraphics[width=.96\textwidth]{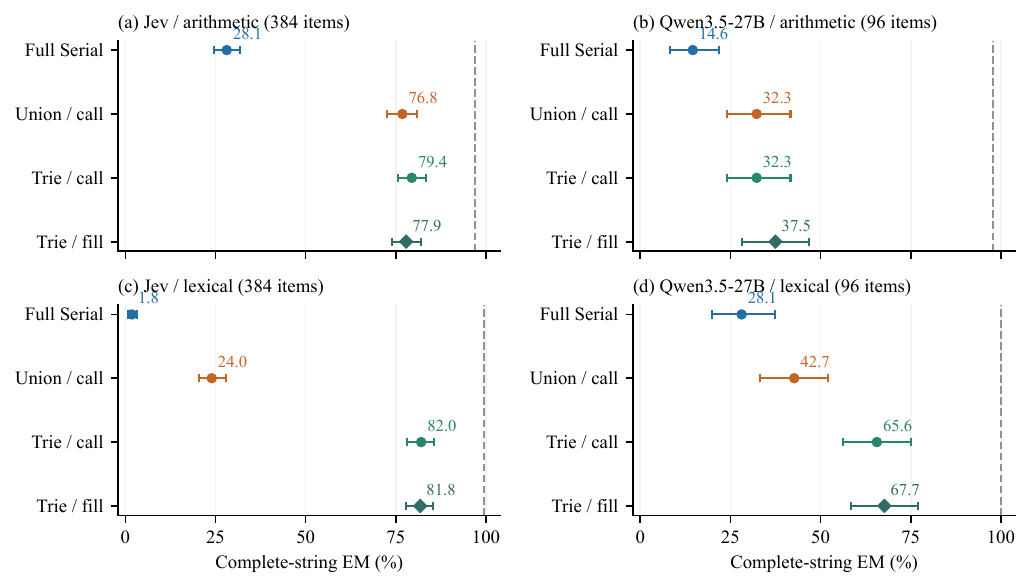}
\caption{Preserving complete candidate consistency restores much of lexical construction. Trie's valid outputs belong to Native's eight candidates; Call still queries singletons, while Fill supplies them. Jev's domain contrast shrinks, but all local conditions remain below Native (dashed reference). Panels use 384 Jev or 96 Qwen parents per domain; intervals resample whole items within length. Native and Full are fresh, interleaved reruns on the same parents; Table~\ref{tab:primary} reports the original responses from the primary collection.}
\label{fig:matched-feasible-outputs}
\end{figure*}

\subsection{Crossed controller factors on longer strings}\label{controller-factor-study}
Thirty-two previously unused twelve-character sources over a--j are
reversed. Each source has two eight-candidate presentations with the
same four-by-two branching and Hamming geometry: branches occur at
positions 2/3 or 10/11. Distractor symbols are sampled from a--j;
all remaining positions agree across candidates. The local alphabet
(a--j or a--z) and number of queried positions (two or six) are crossed
with branch placement. Both branch positions are always queried;
six-query controllers also query positions 1,4,8,12. Other positions
are filled only from singleton candidate unions. These fills use no
gold answer. Query count varies exposure at unambiguous positions while keeping
ambiguous branch count fixed.

Native is shared across the four local arms of each presentation.
All requests retain the original problem, candidates, Serial instruction
and actual prefix. Seed 20261011 freezes items and interleaves factors
within position rounds. Each model makes 1,088 requests. All eight
configurations remain grouped by source in 9,999 bootstrap draws.
Three average factor contrasts per model share six-test Holm correction;
exact sign-permutation distributions use source-level differences.
This fixed-vocabulary reversal task isolates controller factors.

\begin{figure*}[tbp]
\centering
\includegraphics[width=\textwidth]{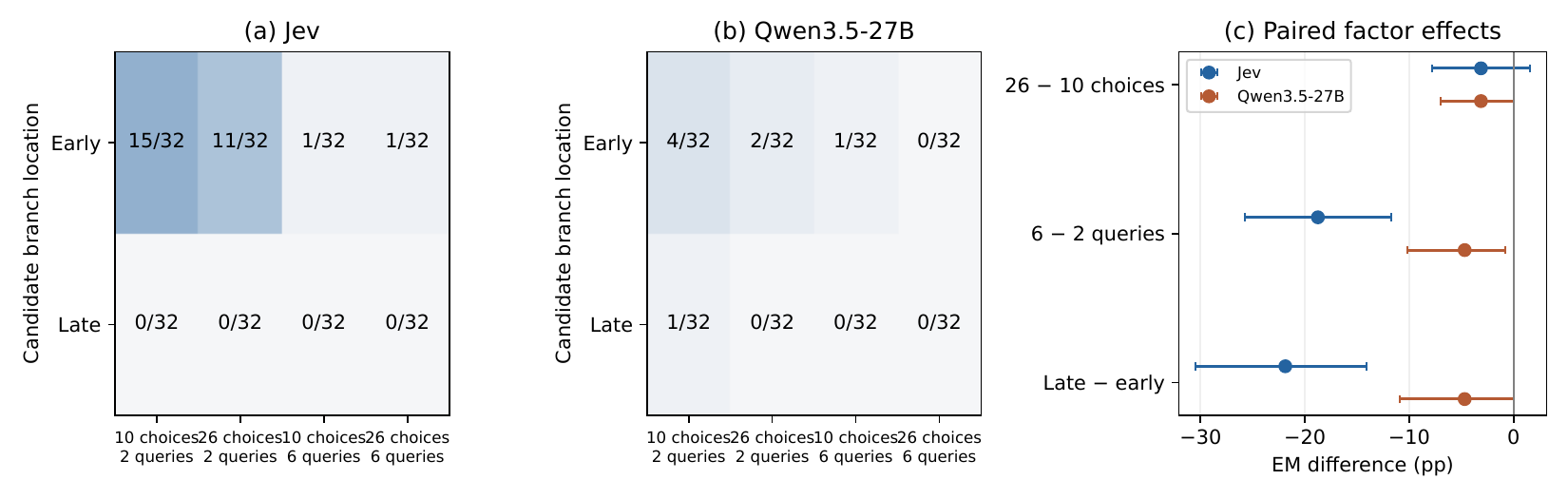}
\caption{Query count and branch placement affect Jev more than alphabet size in this symbolic panel. On 32 twelve-character reversal sources, (a,b) counts show correct strings under crossed factors. (c) Paired effects have 95\% source-bootstrap intervals; Jev's query-count and branch-location contrasts survive six-test Holm correction.}
\label{fig:controller-factors}
\end{figure*}

\begin{table*}[tbp]
\centering\small
\caption{Correctness depends on branch location and queried depth despite a fixed eight-candidate set. Each cell contains 32 sources; counts are complete-output correct and joint/Native correct. Query counts exclude deterministic fills. Exact 95\% upper SCR limits for zero cells are 13.2\% (0/26), 16.1\% (0/21) and 20.6\% (0/16).}
\label{tab:controller-factors}
\setlength{\tabcolsep}{5pt}
\begin{tabular}{@{}llrrrr@{}}
\toprule
Branch & Choices / queries & Jev correct & Jev SCR count & Qwen correct & Qwen SCR count\\
\midrule
Early & 10 / 2 & 15/32 & 15/31 & 4/32 & 3/21 \\
Early & 26 / 2 & 11/32 & 11/31 & 2/32 & 1/21 \\
Early & 10 / 6 & 1/32 & 1/31 & 1/32 & 1/21 \\
Early & 26 / 6 & 1/32 & 1/31 & 0/32 & 0/21 \\
Late & 10 / 2 & 0/32 & 0/26 & 1/32 & 1/16 \\
Late & 26 / 2 & 0/32 & 0/26 & 0/32 & 0/16 \\
Late & 10 / 6 & 0/32 & 0/26 & 0/32 & 0/16 \\
Late & 26 / 6 & 0/32 & 0/26 & 0/32 & 0/16 \\
\bottomrule
\end{tabular}
\end{table*}

\subsection{Solving through continuous and reconstructed-prefix outputs}
\label{natural-generation-interface-control}

On the same 192 Qwen parents, WholeGenerate solves from the original
problem, candidates and length in one uninterrupted response; no answer
is supplied. PrefixContinue removes position/prefix-state fields,
appends its constructed prefix to the assistant prompt, generates a
suffix and consumes its first character. WholeGenerate uses JSON-text
state; PrefixContinue uses structured state. Both allow 32 new tokens,
versus eight for positional queries, with disabled thinking and greedy
decoding. Logs preserve every generated suffix \mbox{and input/output token sequence.}

PrefixContinue is a repeated reconstruction controller, distinct from
continuous decoding. Its initial complete suffix is correct on 94/96
arithmetic and 92/96 lexical items. Invalid or empty subsequent
characters fail and enter the next state as an invalid marker. Zero
lexical completions have an exact 95\% upper limit of 3.8\%.
Six paired generation contrasts have a separate Holm family.

Table~\ref{tab:interface-generation} in the main text reports the paired
content, format and request outcomes.

\subsection{A length control for indexed copying}
\label{indexed-copy-length-control}

Ninety-six parents (eight per domain/length) receive RawCopy,
IndexedCopy and PaddedCopy. Each supplies the gold target and its own
prefix over $L$ queries; question, options and order are fixed.
Nonindexed punctuation matches Indexed's serialized-state character
and UTF-8 byte length at the same prefix. Four Indexed-minus-Padded
contrasts share Holm correction.

A separate Qwen token-length control uses the same parents and three
arms. Punctuation, excluding answer symbols and index numbers, matches
the pinned tokenizer's full rendered Indexed request at the same prefix.
Every realized match is verified; generated prefixes remain arm-specific.
Its two contrasts form a separate Holm family. Whole-item intervals
are stratified by length. Both controls change information and salience
as well as addressing, so they do not isolate an internal mechanism.

\begin{table*}[tbp]
\centering\small
\caption{Indexed copying improves fidelity over nonindexed length controls. Byte matching fixes serialized-state length; token matching uses Qwen's rendered prompt at a common prefix. Each model/domain has 48 parents. Paired differences have 95\% item intervals, with separate Holm correction for byte and token comparisons.}
\label{tab:interface-padding}
\setlength{\tabcolsep}{4pt}
\begin{tabular}{@{}lllrrrr@{}}
\toprule
Model & Domain & Match & Indexed & Padded & Difference [95\% CI] & Holm $p$\\
\midrule
Jev & math & bytes & 43/48 & 17/48 & 54.2 [43.8, 64.6] & 8.94e-08 \\
Jev & lexical & bytes & 47/48 & 8/48 & 81.2 [72.9, 89.6] & 1.46e-11 \\
Qwen3.5-27B & math & bytes & 23/48 & 8/48 & 31.2 [18.8, 43.8] & 0.00146 \\
Qwen3.5-27B & lexical & bytes & 47/48 & 38/48 & 18.8 [10.4, 27.1] & 0.00391 \\
Qwen3.5-27B & math & tokens & 23/48 & 9/48 & 29.2 [16.7, 41.7] & 0.00258 \\
Qwen3.5-27B & lexical & tokens & 47/48 & 34/48 & 27.1 [16.7, 37.5] & 0.000488 \\
\bottomrule
\end{tabular}
\end{table*}

\subsection{An empirical reference for composing imperfect decisions}
\label{empirical-composition-reference}

For each model/domain/length and matched population, let $c_{it}$
indicate position correctness. The empirical independence reference is
$R=\prod_t\overline c_t$, using binary observed outcomes, not reported
model probabilities. Independent uses 384 parents per domain and
TeacherForced 60, with \mbox{all-items and Native-correct populations.}

The reference condition's EM minus $R$ reflects error dependence and
item heterogeneity. Serial minus that condition also changes prefix
context and instructions. In 9,999 paired draws within length, all
position means, products, EM and conditional denominators are
recomputed together. The figure shows Jev; all original models are
included in \mbox{the accompanying numerical results.}

\begin{figure*}[tbp]
\centering
\includegraphics[width=.98\textwidth]{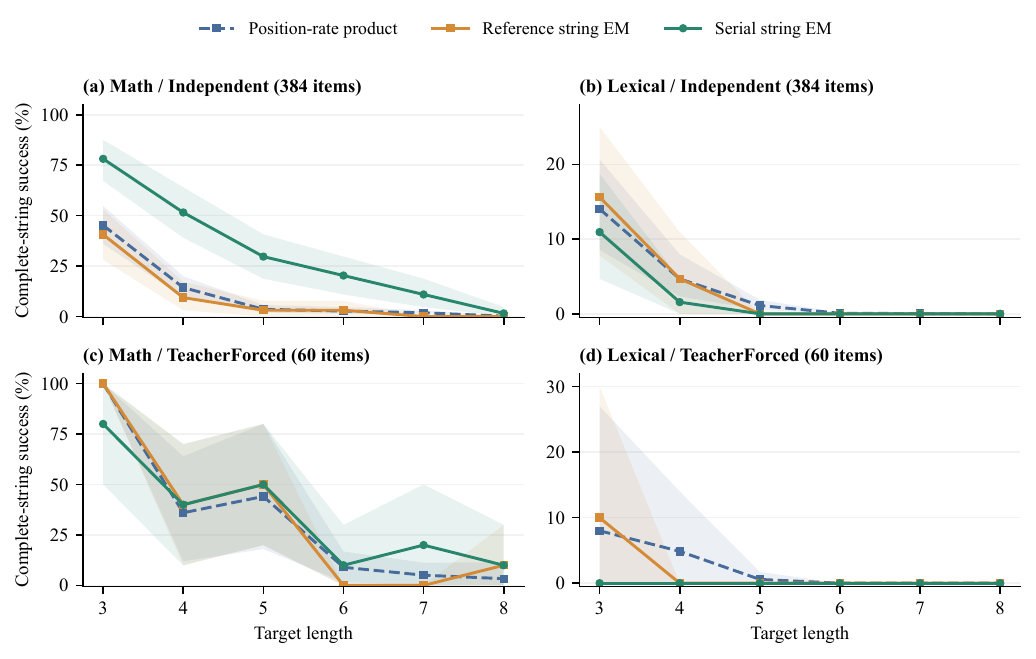}
\caption{Local error rates already imply few correct lexical strings under an independence reference. Jev arithmetic Serial exceeds the Independent reference at most lengths, whereas lexical success remains low in both conditions. TeacherForced comparisons use their smaller matched population. Bands are 95\% whole-item bootstrap intervals; the product is an empirical reference, not a calibrated sequence probability.}
\label{fig:composition-reference}
\end{figure*}

\subsection{Copying with fixed position criteria}
\label{fixed-criterion-copy-follow-up}

Ninety-six parents (eight arithmetic/lexical pairs per length) receive
the same supplied target, absolute-position question and option order.
Raw gives the string; Indexed adds its position table. Independent
omits the draft; Serial supplies its actual prefix. Batch bundles
these questions: native multi-field decisions for Jev/StartLux, strict
JSON for Qwen. Single-position generation allows eight tokens;
malformed output fails without trimming, extraction or case repair.
The nested binding probe changes literal to opaque names at one
preselected position, reusing identical literal requests.

Six Serial indexing and six raw Batch-minus-Independent tests share
Holm correction. Feedback contrasts and twelve raw/indexed binding
contrasts have separate families. Intervals use 99,999 whole-item draws
within length, seed 20261008.

\begin{figure*}[tbp]
\centering
\includegraphics[width=.98\textwidth]{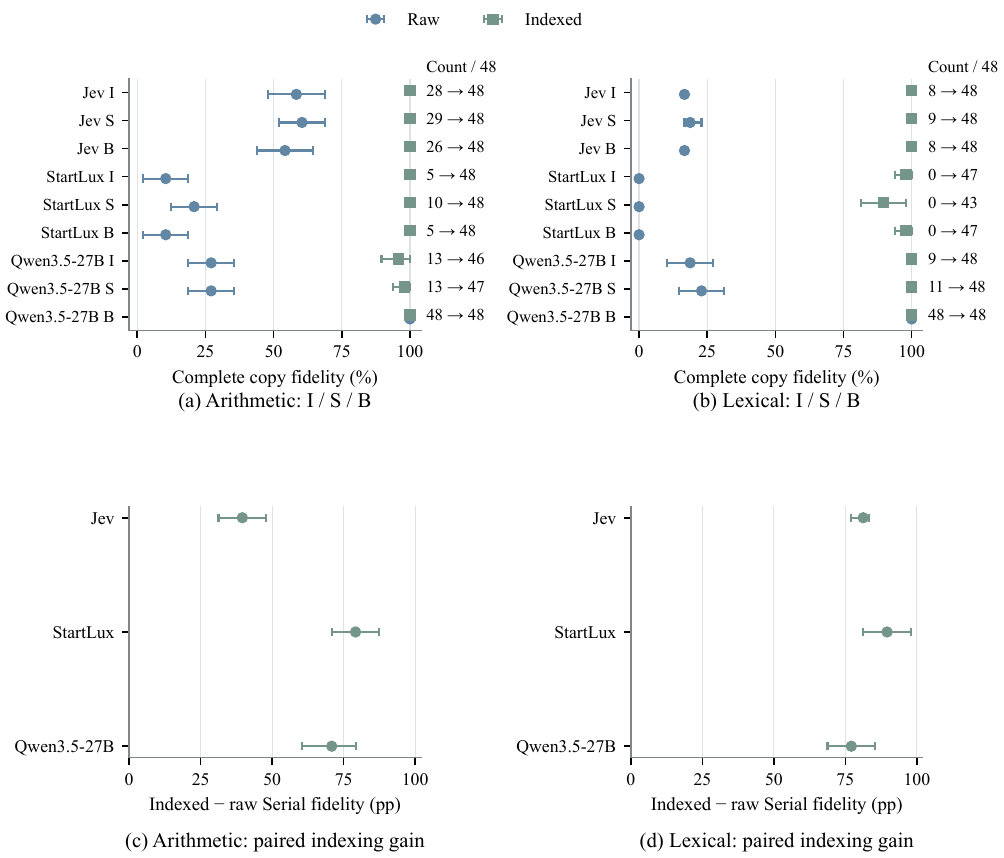}
\caption{Indexing improves complete copying under fixed position questions. With supplied gold targets and 48 parents per domain, (a,b) indexed fidelity is high across controllers, while raw Batch helps Qwen3.5-27B. (c,d) Serial indexing gains are positive for every model/domain. I/S/B denote Independent/Serial/Batch; paired 95\% item bootstrap intervals preserve all conditions within each parent.}
\label{fig:copy-controls}\label{tab:revision5}\label{tab:revision6}
\end{figure*}

\begin{table*}[tbp]
\centering
\small
\caption{Literal names improve Qwen raw readout; indexed readout is correct under both bindings. Each domain has 48 parents. Differences are percentage points with paired 95\% intervals and twelve-test Holm correction.}
\label{tab:revision7}
\setlength{\tabcolsep}{3pt}
\begin{tabular}{@{}llrrrrr@{}}
\toprule
Model & Domain & Evidence & Literal & Opaque & Difference {[}95\% CI{]} & Holm p \\
\midrule
Jev & lexical & raw & 35 & 34 & 2.1 {[}-8.3, 12.5{]} & 1 \\
Jev & math & raw & 43 & 43 & 0.0 {[}-6.2, 6.2{]} & 1 \\
Qwen3.5-27B & lexical & raw & 37 & 11 & 54.2 {[}41.7, 66.7{]} & 3.3e-07 \\
Qwen3.5-27B & math & raw & 41 & 14 & 56.2 {[}41.7, 68.8{]} & 1.8e-07 \\
StartLux & lexical & raw & 11 & 11 & 0.0 {[}0.0, 0.0{]} & 1 \\
StartLux & math & raw & 26 & 26 & 0.0 {[}0.0, 0.0{]} & 1 \\
\bottomrule
\end{tabular}
\end{table*}

\subsection{Copying supplied incorrect targets}
\label{copying-supplied-incorrect-targets}

Ninety-six parents (eight per domain/length) each receive matched
correct/incorrect targets under Raw and Indexed. Incorrect plans comprise
48 observed arithmetic selections, 28 observed lexical selections and
20 constructed lexical candidates; observed errors can come from another
model, and two lexical plans are Jev's own errors. This error-enriched
population measures conditional fidelity, separate from fresh-plan
success prevalence. Records retain difference counts and distinguish observed from
constructed plans.

Six wrong-plan indexing and six indexed wrong-minus-correct tests use
separate Holm families. Whole-parent bootstrap preserves target/evidence
pairs (99,999 draws, seed 20261008). Exact binomial intervals supplement
boundary cells: for example, 48/48 gives [92.6\%,100\%].

\begin{figure*}[tbp]
\centering
\includegraphics[width=.98\textwidth]{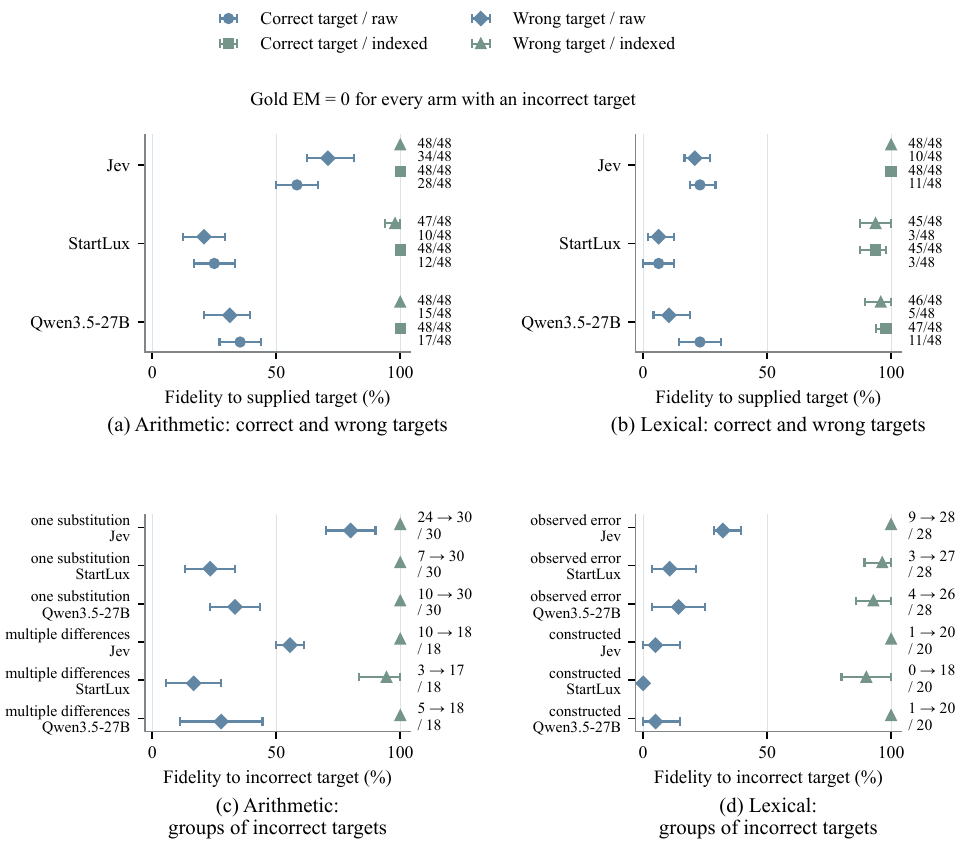}
\caption{Faithful copying can preserve errors. On 48 parents per domain with paired correct/incorrect targets, (a,b) indexing raises fidelity; Jev copies every indexed incorrect target. Gains cover (c) single/multiple arithmetic character differences and (d) observed/constructed lexical errors. Intervals use 95\% item bootstrap.}
\label{fig:wrong-plan-copying}\label{tab:revision12}\label{tab:revision13}
\end{figure*}

\subsection{Crossed prompt and option order robustness}
\label{crossed-prompt-and-option-order-robustness}

Each Qwen model receives 3,456 requests on the same 96 Copy parents:
Indexed/Direct evidence, code/character output, three templates and
three orders at one preselected position. Direct also supplies the
queried character. Both contracts use opaque names, disabled thinking,
greedy decoding and eight tokens. Display permutations preserve code-
character bindings. Malformed responses fail strict accuracy; format-
conditioned accuracy is secondary. Seventy-two setting contrasts share
Holm correction; bootstrap uses 99,999 parent draws within length,
seed 20261008.

\begin{figure*}[tbp]
\centering
\includegraphics[width=.98\textwidth]{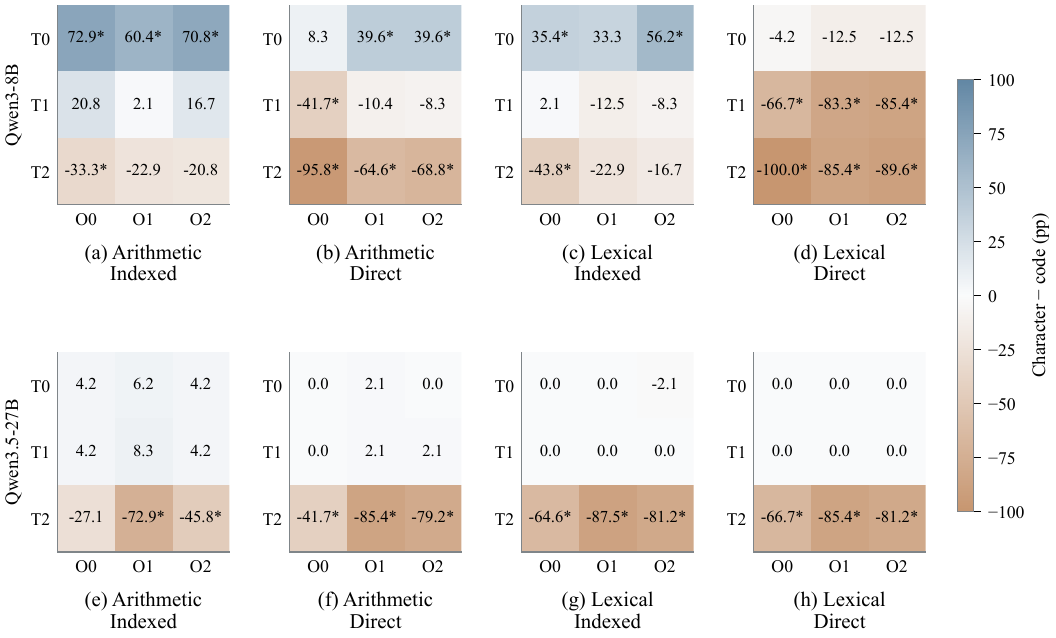}
\caption{Prompt and option order change which output contract is more accurate. For single-position readout on 96 supplied-target parents, (a--d) Qwen3-8B has corrected code and character advantages in different settings; (e--h) Qwen3.5-27B's corrected differences favor codes. Malformed outputs fail strict scoring. Asterisks mark Holm $p<0.05$ after correction across all 72 paired comparisons.}
\label{fig:robustness}
\end{figure*}

\label{template-interactions-and-observable-failure-types}

Template interactions average character-minus-code differences over
orders within parent, then subtract T0 from T1 or T2. Sixteen sign-
randomization tests form a separate Holm family; whole-parent intervals
use seed 20261008 and 99,999 draws. Differences of differences can exceed
100 percentage points. The strict failure taxonomy distinguishes
valid wrong content, bound-code substitutions, other codes, extra text,
invalid output and truncation. Appendix~\ref{app:prompts} reproduces the complete contracts and output examples.

\begin{figure*}[tbp]
\centering
\includegraphics[width=.98\textwidth]{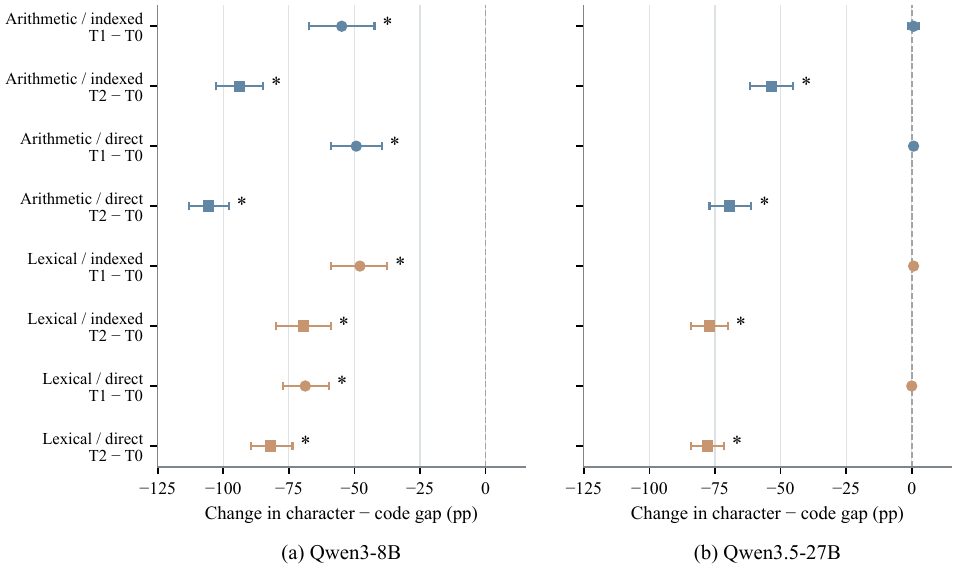}
\caption{Wording shifts the relative success of code and character output. Contrasts average option orders within 96 supplied-target parents. (a) Both alternative templates shift Qwen3-8B's relative accuracy toward codes compared with T0; (b) Qwen3.5-27B changes survive correction for T2 but not T1. Intervals are paired 95\% parent bootstrap; asterisks mark Holm $p<0.05$ after correction across all sixteen interaction tests.}
\label{fig:template-interactions}\label{tab:revision9}
\end{figure*}

\paragraph{Output scoring.}\label{failure-taxonomy-and-interaction-inference}

Scoring priority is strict correct, truncated, empty, format-valid wrong,
correct bound code, another listed code, extra text, then other invalid
output. Codes follow the actual request; whitespace and truncation flags
are also retained. All malformed responses fail. Interaction inference
is specified in Appendix~\ref{template-interactions-and-observable-failure-types}.

\subsection{Calibration and discrimination uncertainty}
\label{calibration-and-discrimination-uncertainty}

The original confidence population has 384 Serial items per domain.
Reported masses, including rounding, are not renormalized. Product and
geometric mean follow emitted characters; minimum top-one probability
can differ when service selections disagree with argmax. Ten fixed
width bins define ECE. Bootstrap uses 99,999 item draws within length,
seed 20261005. Discrimination is undefined in one-class samples;
empty reliability bins remain empty. No calibrator is fitted.

\begin{figure*}[tbp]
\centering
\includegraphics[width=.96\textwidth]{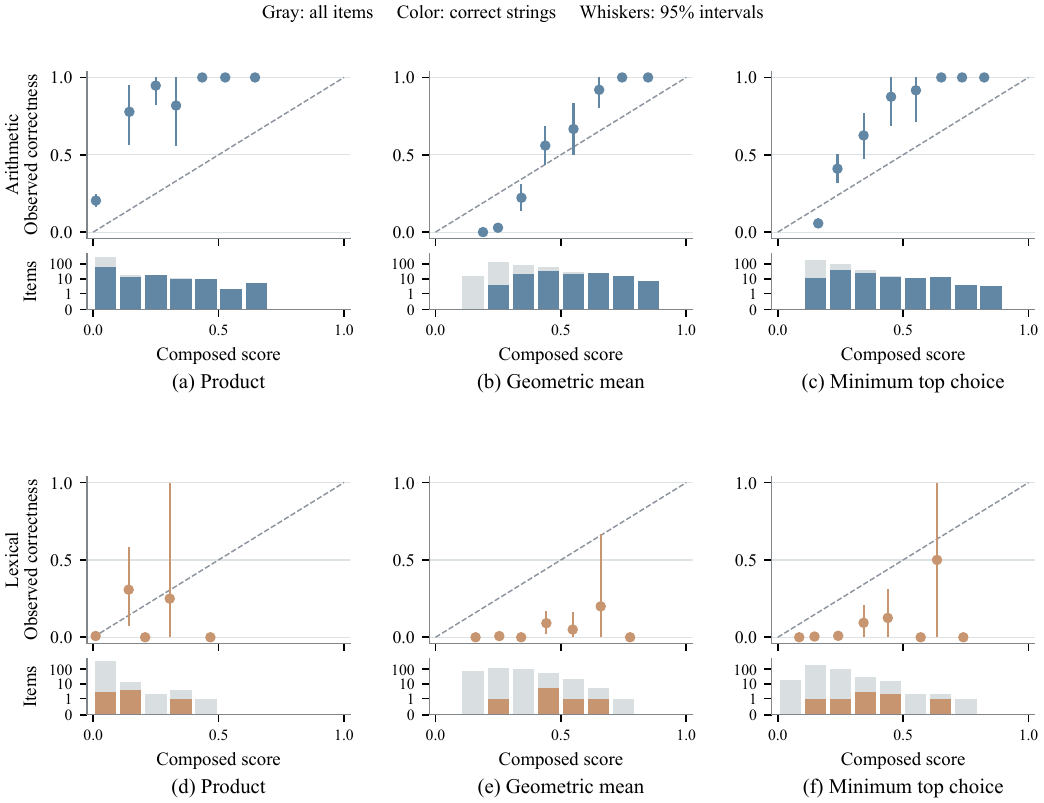}
\caption{High local confidence can coexist with incorrect strings. On 384 Jev items per domain, (a--c) arithmetic scores track correctness but remain miscalibrated; (d--f) lexical successes are sparse, and local-confidence summaries often overstate correctness. Bin counts give each estimate's support. Whole items are resampled within length.}
\label{fig:reliability}
\end{figure*}

\subsection{Independent lexical calibration study}
\label{independent-lexical-calibration-follow-up}

All 2,347 retained independent targets use the original Native/Serial
prompts and pinned Jev. Source quotas and review exclusions precede
outcomes (Appendix~\ref{lexical-sources-and-review}). Every returned
choice is scored, including probability-argmax disagreements.
Bootstrap uses 9,999 draws within length, seed 20261008.
Conditional exact 95\% SCR upper limits are 0.67\% (0/546) and 0.79\%
(0/468) for lengths seven and eight. Thirteen successes overall make
fine-grained selection estimates imprecise.

\begin{figure*}[tbp]
\centering
\includegraphics[width=.98\textwidth]{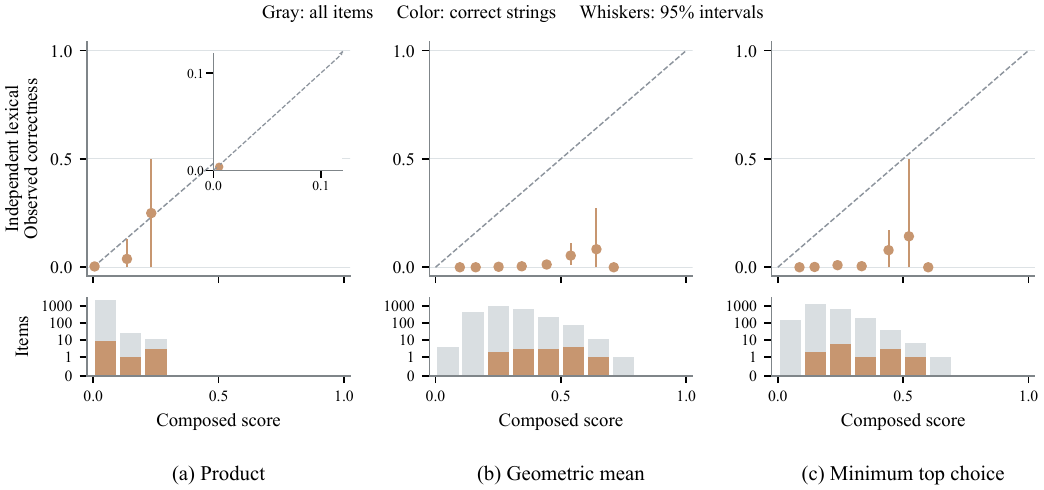}
\caption{Independent lexical evaluation again separates local confidence from string success. Among 2,347 Jev items with 13 Serial successes, (a) Product concentrates near zero; (b,c) higher geometric mean and minimum top choice scores overstate observed correctness. Bins show sparse successes; intervals resample items within length.}
\label{fig:independent-calibration}
\end{figure*}

\begin{table*}[tbp]
\centering
\small
\caption{Native accuracy stays high while independent lexical Serial successes concentrate at shorter lengths. Item bootstrap 95\% intervals stratify by length and redraw the Native denominator. Zero-success cells (\textdagger{}) use exact two-sided binomial intervals in place of degenerate bootstrap bounds.}
\label{tab:revision10}
\setlength{\tabcolsep}{3pt}
\begin{tabular}{@{}llrrrr@{}}
\toprule
Length & Items & Native correct & Serial correct & Retained & SCR (\%) \\
\midrule
3 & 61 & 61 & 7 & 7/61 & 11.48 {[}4.92, 19.67{]} \\
4 & 315 & 310 & 3 & 3/310 & 0.97 {[}0.00, 2.26{]} \\
5 & 442 & 439 & 2 & 2/439 & 0.46 {[}0.00, 1.14{]} \\
6 & 511 & 504 & 1 & 1/504 & 0.20 {[}0.00, 0.60{]} \\
7 & 548 & 546 & 0 & 0/546 & 0.00 {[}0.00, 0.67{]}\textdagger{} \\
8 & 470 & 468 & 0 & 0/468 & 0.00 {[}0.00, 0.79{]}\textdagger{} \\
All & 2347 & 2328 & 13 & 13/2328 & 0.56 {[}0.26, 0.86{]} \\
\bottomrule
\end{tabular}
\end{table*}

\label{ranking-information-beyond-target-length}

Negative target length is a fixed ranking baseline, without Brier/ECE.
Scores share outcomes and bootstrap draws; one-class replicates are
undefined. Source-mixture and equal-length weighting are evaluated
separately. The independent lexical score-minus-length intervals span
zero, so pooled discrimination does not establish incremental screening
utility beyond length.

\begin{table*}[tbp]
\centering\small
\caption{Product ranks arithmetic outcomes better than target length; independent source-mixture lexical gains remain uncertain. Differences compare Product with negative-length ranking, with paired 95\% item intervals.}
\label{tab:revision16}\label{fig:confidence-ranking}
\setlength{\tabcolsep}{5pt}
\begin{tabular}{@{}lrr@{}}
\toprule
Population / mixture & AUROC difference & AP difference\\
\midrule
Arithmetic & 0.068 [0.051, 0.086] & 0.175 [0.135, 0.216] \\
Original lexical & 0.032 [-0.034, 0.085] & 0.128 [0.016, 0.425] \\
New lexical / source & 0.011 [-0.043, 0.064] & 0.055 [-0.012, 0.249] \\
New lexical / equal & 0.032 [-0.013, 0.075] & 0.141 [0.023, 0.444] \\
\bottomrule
\end{tabular}
\end{table*}

\label{common-length-standardization-and-fixed-coverage-selection}

Source quotas reflect the limited short-noun pool. Equal-length
standardization gives each length weight $1/6$ and each item
$1/(6n_L)$; success classes are never rebalanced. Every bootstrap draw
recomputes weighted prevalence, SCR, Brier, ECE, AUROC and AP.
The prevalence reference uses identical weights; cohorts are not paired
item by item. Fixed-coverage precision uses uniform expected inclusion
for cutoff ties and reports its uncertainty. Ranking, calibration and selection utility are evaluated separately
at the observed prevalence.

\begin{figure*}[tbp]
\centering
\includegraphics[width=.98\textwidth]{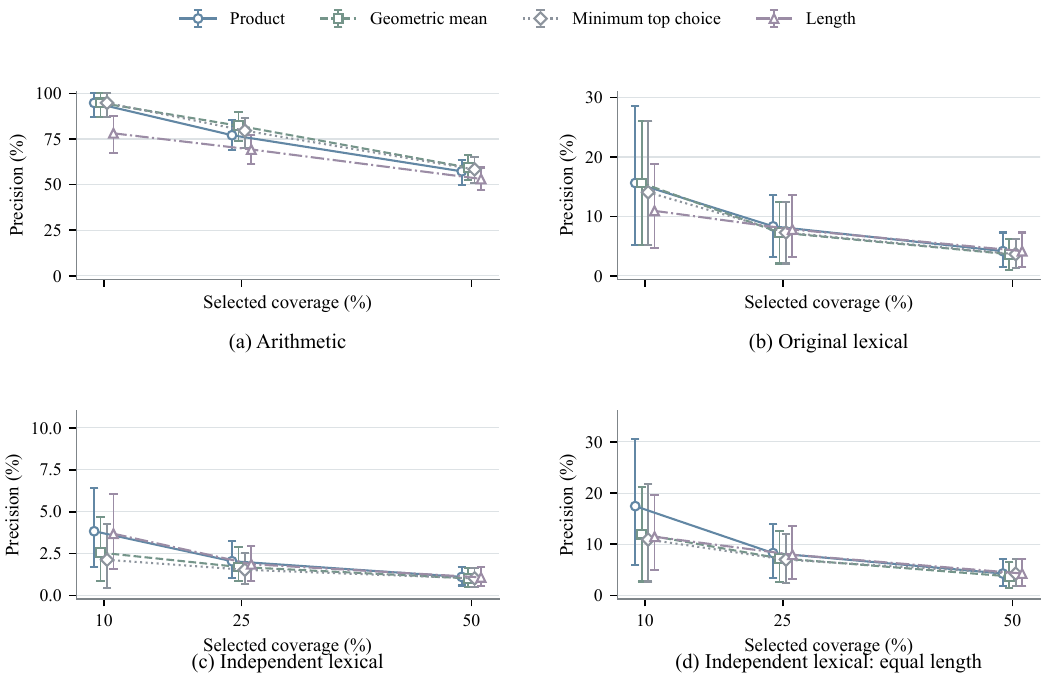}
\caption{Screening precision varies by task and length. (a) Arithmetic precision is high at low coverage; (b) original lexical estimates are imprecise. (c,d) Equal-length weighting raises independent-cohort precision without more successes. Paired 95\% item bootstrap intervals use uniform expected inclusion for cutoff ties.}
\label{fig:selection}
\end{figure*}

\begin{table*}[tbp]
\centering
\small
\caption{Equal-length weighting narrows the lexical SCR difference between cohorts. Accuracy and SCR (\%) have 95\% item bootstrap intervals. Equal length weights preserve each cohort's vocabulary and candidate difficulty.}
\label{tab:revision17}
\setlength{\tabcolsep}{3pt}
\begin{tabular}{@{}llrrr@{}}
\toprule
Population & Mixture & Native EM & Serial EM & SCR \\
\midrule
Original lexical & Equal length & 99.5 {[}98.7, 100.0{]} & 2.1 {[}0.8, 3.6{]} & 2.1 {[}0.8, 3.6{]} \\
New lexical & Source & 99.2 {[}98.8, 99.5{]} & 0.6 {[}0.3, 0.9{]} & 0.6 {[}0.3, 0.9{]} \\
New lexical & Equal length & 99.3 {[}98.9, 99.6{]} & 2.2 {[}0.9, 3.6{]} & 2.2 {[}0.9, 3.6{]} \\
\bottomrule
\end{tabular}
\end{table*}

\begin{table*}[tbp]
\centering
\small
\caption{Lexical calibration and ranking vary with score and length mixture. Intervals are paired 95\% item bootstrap. Equal assigns weight $1/6$ to each length. Geom. denotes geometric mean.}
\label{tab:revision20}
\setlength{\tabcolsep}{3pt}
\begin{tabular}{@{}llrrrrr@{}}
\toprule
Population & Mixture & Score & Brier & ECE & AUROC & AP \\
\midrule
Original & Equal & Product & 0.019 {[}0.009, 0.030{]} & 0.010 {[}0.005, 0.025{]} & 0.936 {[}0.877, 0.978{]} & 0.231 {[}0.081, 0.543{]} \\
Original & Equal & Geom. & 0.109 {[}0.101, 0.117{]} & 0.286 {[}0.269, 0.301{]} & 0.849 {[}0.659, 0.959{]} & 0.164 {[}0.045, 0.461{]} \\
Original & Equal & Minimum & 0.057 {[}0.050, 0.064{]} & 0.183 {[}0.169, 0.198{]} & 0.871 {[}0.724, 0.966{]} & 0.174 {[}0.048, 0.477{]} \\
New & Source & Product & 0.005 {[}0.003, 0.008{]} & 0.003 {[}0.001, 0.006{]} & 0.910 {[}0.830, 0.973{]} & 0.126 {[}0.041, 0.351{]} \\
New & Source & Geom. & 0.096 {[}0.093, 0.098{]} & 0.286 {[}0.281, 0.290{]} & 0.838 {[}0.706, 0.942{]} & 0.066 {[}0.020, 0.191{]} \\
New & Source & Minimum & 0.044 {[}0.042, 0.046{]} & 0.184 {[}0.180, 0.187{]} & 0.806 {[}0.680, 0.909{]} & 0.049 {[}0.012, 0.187{]} \\
New & Equal & Product & 0.019 {[}0.009, 0.030{]} & 0.008 {[}0.003, 0.021{]} & 0.926 {[}0.864, 0.965{]} & 0.248 {[}0.089, 0.568{]} \\
New & Equal & Geom. & 0.107 {[}0.102, 0.114{]} & 0.283 {[}0.269, 0.296{]} & 0.857 {[}0.726, 0.946{]} & 0.175 {[}0.051, 0.434{]} \\
New & Equal & Minimum & 0.056 {[}0.051, 0.063{]} & 0.181 {[}0.168, 0.194{]} & 0.859 {[}0.763, 0.933{]} & 0.161 {[}0.043, 0.437{]} \\
\bottomrule
\end{tabular}
\end{table*}

\subsection{Paired correctness turnover}\label{paired-correctness-turnover}
Table~\ref{tab:paired-diagnostics} reconstructs the four joint outcomes
from paired responses. SCR conditions on Native success; Serial success
on Native-wrong items describes gains outside that population. The
natural-task model ordering agrees under SCR and Serial EM. Semantic
fields have lower Native coverage and larger conditional differences:
Qwen named-field accuracy gains coexist with substantial losses of
previously correct items. Marginal accuracies alone leave this overlap
unidentified; its mathematical bounds appear in
Appendix~\ref{paired-outcome-interpretation}.

\begin{table*}[!t]
\centering\small
\caption{Marginal improvement can coexist with substantial correctness turnover. Joint counts give both correct / Native only / Serial only / neither. SCR, its difference from Serial EM ($\Delta$, percentage points), and Serial success on Native-wrong items (Gain rate) summarize different parts of the same paired table.}
\label{tab:paired-diagnostics}
\setlength{\tabcolsep}{5pt}
\begin{tabular}{@{}llrrrrr@{}}
\toprule
Model / interface & Task & Items & Joint counts & SCR (\%) & $\Delta$ & Gain rate (\%)\\
\midrule
Jev & lexical & 384 & 8/374/0/2 & 2.1 & +0.0 & 0.0 \\
Jev & math & 384 & 123/250/0/11 & 33.0 & +0.9 & 0.0 \\
Qwen3-8B & lexical & 384 & 0/363/0/21 & 0.0 & +0.0 & 0.0 \\
Qwen3-8B & math & 384 & 0/59/0/325 & 0.0 & +0.0 & 0.0 \\
Qwen3.5 / code & lexical & 384 & 5/378/0/1 & 1.3 & +0.0 & 0.0 \\
Qwen3.5 / code & math & 384 & 1/96/1/286 & 1.0 & +0.5 & 0.3 \\
Qwen3.5 / direct & lexical & 384 & 132/248/0/4 & 34.7 & +0.4 & 0.0 \\
Qwen3.5 / direct & math & 384 & 39/339/0/6 & 10.3 & +0.2 & 0.0 \\
StartLux & lexical & 384 & 1/359/0/24 & 0.3 & +0.0 & 0.0 \\
StartLux & math & 384 & 8/227/1/148 & 3.4 & +1.1 & 0.7 \\
Jev & binary fields & 128 & 80/7/29/12 & 92.0 & +6.8 & 70.7 \\
Jev & named fields & 128 & 111/4/5/8 & 96.5 & +5.9 & 38.5 \\
Qwen3-8B & binary fields & 128 & 1/18/19/90 & 5.3 & -10.4 & 17.4 \\
Qwen3-8B & named fields & 128 & 6/22/31/69 & 21.4 & -7.5 & 31.0 \\
StartLux & binary fields & 128 & 25/25/13/65 & 50.0 & +20.3 & 16.7 \\
StartLux & named fields & 128 & 51/15/40/22 & 77.3 & +6.2 & 64.5 \\
\bottomrule
\end{tabular}

\par\medskip
\caption{Most Full lexical outputs are absent from WordNet, while Trie completions stay within the candidates. Counts list gold / other candidate / malformed / other noun / other word / unindexed. No gold-synset alternatives were found. Recombinations are noncandidate strings compatible with every position-wise union; they can overlap the resource categories. This is an automated resource audit, not semantic adjudication.}
\label{tab:lexical-output-audit}
\setlength{\tabcolsep}{6pt}
\begin{tabular}{@{}llrrr@{}}
\toprule
Model & Controller & Items & Output counts & Recombinations\\
\midrule
Jev & Full & 384 & 7/1/0/11/0/365 & 22 \\
Jev & TrieCall & 384 & 315/69/0/0/0/0 & 0 \\
Jev & TrieFill & 384 & 314/70/0/0/0/0 & 0 \\
Jev & UnionCall & 384 & 92/9/0/11/1/271 & 283 \\
Qwen3.5-27B & Full & 96 & 27/7/0/2/1/59 & 2 \\
Qwen3.5-27B & TrieCall & 96 & 63/28/5/0/0/0 & 0 \\
Qwen3.5-27B & TrieFill & 96 & 65/31/0/0/0/0 & 0 \\
Qwen3.5-27B & UnionCall & 96 & 41/14/22/1/0/18 & 19 \\
\bottomrule
\end{tabular}

\par\medskip
\caption{Position-query failure occurs without hitting the output cap. Budgets and observed maxima are new tokens, including special tokens; truncated counts use recorded request flags.}
\label{tab:generation-budget}
\setlength{\tabcolsep}{3pt}
\begin{tabular}{@{}llrrrr@{}}
\toprule
Task & Interface & Requests & Cap & Max & Truncated\\
\midrule
lexical & Position & 528 & 8 & 2 & 0 \\
lexical & PrefixContinue & 528 & 32 & 32 & 4 \\
lexical & WholeGenerate & 96 & 32 & 4 & 0 \\
math & Position & 528 & 8 & 2 & 0 \\
math & PrefixContinue & 528 & 32 & 9 & 0 \\
math & WholeGenerate & 96 & 32 & 9 & 0 \\
\bottomrule
\end{tabular}
\end{table*}

\subsection{Lexical output membership and semantic review}\label{lexical-output-audit}
We classify every lexical output in the matched Full/UnionCall/TrieCall/
TrieFill panel against its candidates and WordNet 3.0. In order, categories
are gold, another candidate, malformed (wrong length or outside lowercase
ASCII), another gold-synset lemma, another indexed noun, another indexed
word, and unindexed string. No off-candidate output is another lemma of
the gold synset. Position-wise candidate recombination is recorded separately from
semantic classification.

Two reviewers independently assessed 50 distinct item/output pairs:
all 26 indexed noncandidate pairs and four unindexed pairs per length,
selected by ascending stable subject ID. Materials showed the definition,
required length and proposed word, withholding gold targets, candidates,
models and controllers. Reviewers judged noun validity and definition
fit separately using yes, no or uncertain. Their judgments agreed on
both dimensions for every pair: 33 outputs were rejected as nouns,
17 were accepted as nouns, and 49 did not fit the definition. The remaining
pair, \emph{pop} for ``a quick short straight punch,'' received uncertain
definition-fit judgments from both reviewers.

No reviewed pair received unanimous noun-and-definition acceptance,
so adding documented semantic alternatives leaves EM unchanged.
Accepting the uncertain pair would change Jev UnionCall from 92/384 to
93/384 (24.0\% to 24.2\%); the other controller results remain unchanged.
The 50 pairs map to 51 observed outputs because one pair recurs across
controllers. This sensitivity calculation covers reviewed pairs;
unindexed outputs outside the sample retain unresolved semantic status.
Table~\ref{tab:lexical-output-audit} reports resource categories separately
from these human judgments. All analyses retain strict closed-set EM as the task score.

\subsection{Observed generation budgets}\label{generation-budget-audit}
Raw successful request records retain generated token IDs and the
truncation flag. All 1,056 position-query requests stop within two tokens,
below the eight-token cap (Table~\ref{tab:generation-budget}). The 32-token
whole-generation cap also never binds. Four lexical PrefixContinue
requests reach their cap; its complete-trajectory format failures also
include responses below the cap. This audit rules out position-budget
truncation in this panel. It leaves prompt-state representation, assistant
prefill tokenization and repeated requests jointly changed, so the
comparison identifies controller outcomes rather than an isolated
internal autoregressive mechanism.

\FloatBarrier
\setcounter{table}{0}
\renewcommand{\thetable}{E\arabic{table}}
\setcounter{dbltopnumber}{1}
\section{Complete Prompts and Observable Output
Failures}
\label{app:prompts}\label{complete-prompts-and-observable-output-failures}

\subsection{Formal content and stopping
instructions}\label{formal-content-and-stopping-instructions}

Boxes reproduce the request text verbatim. Blue braces denote item-
specific substitutions; they are not literal model input. The content
track queries $L$ absolute positions without END.

\begin{promptbox}{Native}
\raggedright
Select the complete correct \mbox{answer from the provided candidates.}
\end{promptbox}

\begin{promptbox}{Independent}
\raggedright
Solve the original problem. Select its true answer character at the specified one-based \mbox{position, independently of other positions.}
\end{promptbox}

\begin{promptbox}{Serial}
\raggedright
Solve the original problem. Select its true answer character at the specified position. The current prefix is a draft and may contain errors; it \mbox{does not change the original answer.}
\end{promptbox}

TeacherForced shares the Serial instruction and substitutes the gold
prefix in the state below.

\begin{promptbox}{Gold Copy}
\raggedright
Copy copy\_target exactly. Select its character at \mbox{the specified one-based position.}
\end{promptbox}

\begin{table*}[tbp]
\centering
\begin{minipage}[t]{.48\textwidth}
\vspace{0pt}
\begin{promptbox}{Serial / generated-prefix state}
\texttt{candidates:} \promptvar{candidate\_strings}\\
\texttt{current\_prefix:} \promptvar{generated\_prefix}\\
\texttt{position:} \promptvar{one\_based\_position}\\
\texttt{problem:} \promptvar{problem}\\
\texttt{target\_length:} \promptvar{length}
\end{promptbox}
\end{minipage}\hfill
\begin{minipage}[t]{.48\textwidth}
\vspace{0pt}
\begin{promptbox}{TeacherForced / gold-prefix state}
\texttt{candidates:} \promptvar{candidate\_strings}\\
\texttt{current\_prefix:} \promptvar{gold\_prefix}\\
\texttt{position:} \promptvar{one\_based\_position}\\
\texttt{problem:} \promptvar{problem}\\
\texttt{target\_length:} \promptvar{length}
\end{promptbox}
\end{minipage}
\caption{Serial and TeacherForced share instructions and canonical JSON state; only prefix sources differ.}
\label{tab:prefix-prompt-pair}
\end{table*}

Every state contains problem and target\_length. Native additionally
contains all eight candidates. Independent adds the position, indexed from one.
Serial adds position and current\_prefix to the same problem/candidate
state. TeacherForced uses the same text and state keys but substitutes
the gold prefix. Gold Copy contains problem, target\_length, position,
current\_prefix, and copy\_target, and withholds the complete answer
candidates. Original SelfPlanSpell uses the Copy instruction and its
selected plan as copy\_target, retaining candidates. State is serialized
as canonical JSON. Local criteria are all ten digits or all 26 lowercase
letters, with deterministic displayed order. Each request reconstructs
state. Native instead uses the eight complete strings as criteria.

The stopping track permits END among its 51 options and uses
the following Serial instruction:

\begin{promptbox}{Serial with termination}
\raggedright
Solve the original problem. Select the next character after current\_prefix. Select END \mbox{immediately after the complete answer.}
\end{promptbox}

It uses the same local state construction with a maximum of L+3
requests. The alphabet is lowercase a to z, digits 0 to 9, the eleven
punctuation symbols period, comma, exclamation mark, question mark,
semicolon, colon, apostrophe, quotation mark, hyphen, left parenthesis
and right parenthesis, followed by space, newline, tab and END before
deterministic shuffling. These punctuation identities are an explicit
project reconstruction. The content track makes L position queries and excludes END
from its options.

\subsection{Complete natural task paraphrases and candidate
support}\label{complete-natural-task-paraphrases-and-candidate-support}

The original 72 item robustness panel uses v1 to v3 for Native,
Independent and Serial. The Qwen3.5-27B panel uses the same Native and
Serial strings. Natural paraphrases change only the instruction,
retaining the controller's problem, candidates, length, position and
prefix. The T0--T2 readout templates below supply the target answer.

\begin{promptbox}{v2, Native}
Which listed full answer correctly solves the problem?
\end{promptbox}

\begin{promptbox}{v2, Independent}
Which character occupies the specified one-based position of the answer
to the original problem?
\end{promptbox}

\begin{promptbox}{v2, Serial}
Given the possibly incorrect draft prefix, identify the original answer
character at the specified position.
\end{promptbox}

\begin{promptbox}{v3, Native}
\raggedright
Identify the correct \mbox{complete solution among the candidate answers.}
\end{promptbox}

\begin{promptbox}{v3, Independent}
\raggedright
Determine the complete solution and identify the \mbox{character at the specified position.}
\end{promptbox}

\begin{promptbox}{v3, Serial}
Use the draft history to determine the true solution character at this
position; the draft may be wrong.
\end{promptbox}

Position union Serial uses exactly the v1 Serial instruction and full
state above. Its displayed local options are the seeded full list
filtered to characters occurring at that position in any candidate. Their stable identities and relative order are retained. The controller returns a singleton without a request. Both Serial arms use the problem and full candidate set without receiving the fresh Native selection.

For Qwen3.5-27B natural queries, a single user message contains JSON
keys state, question, allowed\_values, and response\_format. State is
the canonical JSON string used by the other interfaces, and question is
the applicable instruction above. The Native response contract is:

\begin{promptbox}{Native direct-output contract}
\raggedright
Output only the requested complete string. \mbox{No explanation or markdown.}
\end{promptbox}

The local response contract is:

\begin{promptbox}{Local direct-output contract}
Output only the single requested lowercase letter or digit.
\end{promptbox}

Requests use a single user message and greedy decoding with thinking disabled. The budgets are 32 new tokens for Native and eight for a local query. Options constrain the prompt without a token mask. After outer whitespace removal, scoring requires an exact listed value and a complete response. Format failures and truncated responses fail without repair.

\subsection{Continuous solving and prefix continuation}
\label{continuous-solving-prompts}
Both generation arms share these instructions:
\begin{promptbox}{Complete-answer solving}
Solve the original problem and output only its complete answer, without
explanation or markdown.
\end{promptbox}
\begin{promptbox}{Complete-answer response contract}
\raggedright
Output only the complete answer. \mbox{No explanation or markdown.}
\end{promptbox}
Both receive the original problem, eight candidates and target length;
neither receives the selected or gold answer. WholeGenerate places
canonical JSON text in the user object's \texttt{state} field.
PrefixContinue places the state object there and appends
\promptvar{generated\_prefix} after the assistant generation boundary.
Its state excludes absolute position and current-prefix fields. Both
allow 32 new tokens; PrefixContinue consumes only the first suffix
character per request. Union/Trie and the controller-factor panel retain
the verbatim Serial instruction, with the option/fill rules in
Appendix~\ref{matched-feasible-output-control}.

\begin{table*}[t]
\centering
\begin{minipage}[t]{.48\textwidth}
\vspace{0pt}
\begin{promptbox}{T0 / Code response}
Return exactly ONE listed choice code, as a single uppercase letter. Do
not output the option name or criterion value. Do not explain, reason
aloud, add whitespace, punctuation, or any other text.
\end{promptbox}
\end{minipage}\hfill
\begin{minipage}[t]{.48\textwidth}
\vspace{0pt}
\begin{promptbox}{T0 / Character response}
Return exactly ONE character described by the correct criterion: a
lowercase letter or digit, as appropriate for the listed criteria. Do
not output a choice code or option name. Do not explain, reason aloud,
add whitespace, punctuation, or any other text.
\end{promptbox}
\end{minipage}
\par\medskip
\begin{minipage}[t]{.48\textwidth}
\vspace{0pt}
\begin{promptbox}{T1 / Code response}
\raggedright
Your entire response must be exactly one uppercase letter: the code of the selected choice. Use its criterion to decide. Omit the option name and criterion value. Include no explanation, reasoning, whitespace, \mbox{punctuation, or additional text.}
\end{promptbox}
\end{minipage}\hfill
\begin{minipage}[t]{.48\textwidth}
\vspace{0pt}
\begin{promptbox}{T1 / Character response}
\raggedright
Your entire response must be exactly one character: the lowercase letter or digit specified by the correct choice's criterion. Omit the code and option name. Include no explanation, reasoning, \mbox{whitespace, punctuation, or additional text.}
\end{promptbox}
\end{minipage}
\par\medskip
\begin{minipage}[t]{.48\textwidth}
\vspace{0pt}
\begin{promptbox}{T2 / Code response}
\raggedright
Choose the option whose criterion answers the question. Reply with its single uppercase choice code alone. Your response must contain no other \mbox{characters, explanation, or formatting.}
\end{promptbox}
\end{minipage}\hfill
\begin{minipage}[t]{.48\textwidth}
\vspace{0pt}
\begin{promptbox}{T2 / Character response}
Choose the option whose criterion answers the question. Reply with the
single lowercase letter or digit given by that criterion alone. Your
response must contain no other characters, explanation, or formatting.
\end{promptbox}
\end{minipage}
\caption{T0--T2 preserve the target, criterion bindings and user question. Verbatim code/character instructions differ in response representation; rows change emphasis on the criterion, choice and response boundary.}
\label{tab:contract-t0}\label{tab:contract-t1}\label{tab:contract-t2}
\end{table*}

\begin{table*}[t]
\centering
\begin{minipage}[t]{.48\textwidth}
\vspace{0pt}
\begin{promptbox}{Indexed readout / correct character (T0)}
\texttt{copy\_target: 7470}\\
\texttt{query\_position: 3}\\
\texttt{indexing: one-based}\\
\texttt{indexed\_characters:}\\
\texttt{[1:7, 2:4, 3:7, 4:0]}\\
Each pair is a position/character object.\\[4pt]
\centering
\begin{tabular}{@{}llr@{}}
\toprule
Code & option\_name & criterion \\
\midrule
A & o\_266aa5886067 & 7 \\
B & o\_391552c099c1 & 1 \\
C & o\_986b79d4b69d & 9 \\
D & o\_a4aab3f1f080 & 3 \\
E & o\_98089e6d36f7 & 0 \\
F & o\_d10a4bc9e0c1 & 5 \\
G & o\_92e9e7e5922d & 6 \\
H & o\_ade1a1c2a61a & 8 \\
I & o\_2bf175f9655e & 4 \\
J & o\_cc11310c456c & 2 \\
\bottomrule
\end{tabular}
\par\raggedright\vspace{4pt}
Complete response: \texttt{"7"}.\\
Required character: \texttt{7}. No truncation.
\end{promptbox}
\end{minipage}\hfill
\begin{minipage}[t]{.48\textwidth}
\vspace{0pt}
\begin{promptbox}{Indexed readout / bound code (T2)}
\texttt{copy\_target: 20936748}\\
\texttt{query\_position: 3}\\
\texttt{indexing: one-based}\\
\texttt{indexed\_characters:}\\
\texttt{[1:2, 2:0, 3:9, 4:3,}\\
\texttt{5:6, 6:7, 7:4, 8:8]}\\
Each pair is a position/character object.\\[4pt]
\centering
\begin{tabular}{@{}llr@{}}
\toprule
Code & option\_name & criterion \\
\midrule
A & o\_ade1a1c2a61a & 8 \\
B & o\_a4aab3f1f080 & 3 \\
C & o\_cc11310c456c & 2 \\
D & o\_266aa5886067 & 7 \\
E & o\_92e9e7e5922d & 6 \\
F & o\_986b79d4b69d & 9 \\
G & o\_2bf175f9655e & 4 \\
H & o\_d10a4bc9e0c1 & 5 \\
I & o\_98089e6d36f7 & 0 \\
J & o\_391552c099c1 & 1 \\
\bottomrule
\end{tabular}
\par\raggedright\vspace{4pt}
Complete response: \texttt{"F"}.\\
Required character: \texttt{9}. No truncation.
\end{promptbox}
\end{minipage}
\caption{Correct criterion identification does not guarantee output compliance. The left response emits the required character; the right emits its bound code, which fails the same character contract.}
\label{tab:prompt1}\label{tab:prompt2}
\end{table*}

\subsection{Matched field construction
contract}\label{matched-field-construction-contract}

Every requested field uses the following instruction, followed by the
literal suffix \texttt{Field:}, the field name, \texttt{.}, and its
schema description:

\begin{promptbox}{Current-state field decision}
Determine the current service state after the latest USER turn. Copy
categorical values exactly from the schema. Preserve earlier
user-provided slot values unless the user changes or cancels them.
Ignore SYSTEM suggestions unless accepted by the user. UNSET means no
value is held; DONTCARE means the user explicitly has no preference.
Select the current active intent, or NONE if no intent is active. Return
each requested field. Field: \promptvar{field\_name}.
\promptvar{schema\_description}
\end{promptbox}

Joint requests all fields. Independent requests only the current field.
Both states contain schema and dialogue\_history. Serial requests the
current field and additionally includes partial\_current\_state
containing its actual preceding field predictions for that turn. The
field order is fixed before outcomes. Criteria use stable value IDs and
the corresponding schema value descriptions. The same field order and
value order are reused across arms.

Jev receives these typed questions directly. Qwen3.5-27B receives one
user message JSON object with state, field\_queries, and
response\_format, with no added system message. The response contract is reproduced below:

\begin{promptbox}{Dialogue JSON response}
Return only one JSON object mapping precisely the requested fields to
their exact semantic values, not option IDs. No explanation or markdown.
\end{promptbox}

All three arms use a cap of 256 new tokens and greedy decoding with
thinking disabled. The response must parse as one JSON object with
precisely the requested keys, each mapped to an exact legal string
value. Extra keys, malformed JSON, illegal values and truncation fail
without repair. This field contract specifies JSON keys and legal values.
The character contract specifies a single symbol.

\subsection{Full crossed readout
contracts}\label{full-crossed-readout-contracts}

T0, T1 and T2 specify the same supplied target and output contract. Their wording emphasizes the criterion, the choice, and the response boundary differently. The grid measures model sensitivity to these wording differences. Each request contains one system message with the
following text and one user message containing ordered JSON. Thinking is
disabled, decoding is ordinary greedy, and the maximum is eight new
tokens for both contracts. Model identities are in Appendix~\ref{models-and-reproducibility}.

Across T0 to T2, response length, allowable symbols and surrounding text
remain fixed. T0 names the correct criterion, T1 the correct choice's
criterion, and T2 asks for a choice before its criterion. Answer and criterion bindings remain fixed.

\begin{promptbox}{Shared user question (all six contracts)}
Select the character of copy\_target at the one-based query\_position.
Auxiliary representations, when present, encode the same target. Each
criterion description specifies an output character. Select by the
criterion description; option names identify criteria.
\end{promptbox}

\begin{promptbox}{Code response\_format}
\raggedright
Select one choice and \mbox{output only its single-character code.}
\end{promptbox}

\newpage
\begin{promptbox}{Character response\_format}
\raggedright
Select one choice and output only the \mbox{single character described by its criterion.}
\end{promptbox}

User JSON has exactly state, question, choices, and response\_format.
All arms supply copy\_target, query\_position, indexing set to
\texttt{one-based}, and the complete indexed\_characters list. Direct evidence
additionally supplies query\_character. Each ordered choices entry maps
a displayed uppercase code to option\_name and criterion. Codes bind to
characters once per parent. O0 to O2 permute display order while
preserving those bindings. Option names, response codes and character criteria are distinct
in the opaque condition.

\subsection{Complete request examples}\label{complete-archived-request-and-output-examples}
The two requests below use indexed arithmetic evidence, character
output and the reference order. Each uses its complete T0 or T2 system
message, the shared user question and character response format above.
State and ordered choices are printed in full; no direct query-character
field is supplied. Thinking is disabled. Decoded responses are scored without case repair, whitespace
trimming or code mapping.

\end{document}